%% file: main.tex
\newcommand{\Guoxin}[1]{{\textcolor{blue}{[Guoxin::~#1]}}}

\newcommand{\rev}[2]{\textcolor{black}{#2}}

\documentclass[acmtog]{acmart}
\usepackage{wrapfig} 
\AtBeginDocument{%
  }

\usepackage{booktabs}
\usepackage{multirow}
\usepackage[normalem]{ulem}

\usepackage[ruled,vlined]{algorithm2e}
\setcopyright{cc}
\setcctype{by}
\acmJournal{TOG}
\acmYear{2025} \acmVolume{44} \acmNumber{6} \acmArticle{} \acmMonth{12} \acmPrice{}\acmDOI{XXXXXXX.XXXXXXX}
\begin{document}

\title{INF-3DP: Implicit Neural Fields for Collision-Free Multi-Axis 3D Printing}

\author{Jiasheng Qu}
\email{jsqu@mae.cuhk.edu.hk}

\affiliation{%
  \institution{The Chinese University of Hong Kong}
  \city{Hong Kong}
  \country{China}
}

\author{Zhuo Huang}
\email{zhuohuang0915@gmail.com}

\affiliation{%
  \institution{The Chinese University of Hong Kong, China / The University of Manchester}
  \city{Manchester}
  \country{United Kingdom}
}

\author{Dezhao Guo}
\email{gdez335@gmail.com}

\affiliation{%
  \institution{The Chinese University of Hong Kong}
  \city{Hong Kong}
  \country{China}
}

\author{Hailin Sun}
\email{sunhailin83@gmail.com}

\affiliation{%
  \institution{The Chinese University of Hong Kong}
  \city{Hong Kong}
  \country{China}
}

\author{Aoran Lyu}
\email{lvaoran@hotmail.com}

\affiliation{%
  \institution{The University of Manchester}
  \city{Manchester}
  \country{United Kingdom}
}

\author{Chengkai Dai}
\email{ckdai@cpii.hk}

\affiliation{%
  \institution{Centre for Perceptual and Interactive Intelligence}
  \city{Hong Kong}
  \country{Hong Kong, China}
}

\author{Yeung Yam}
\email{yyam@mae.cuhk.edu.hk}

\affiliation{%
  \institution{The Chinese University of Hong Kong / Centre for Perceptual and Interactive Intelligence}
  \city{Hong Kong}
  \country{China}
}

\author{Guoxin Fang}
\email{guoxinfang@cuhk.edu.hk}
\authornote{Corresponding author: guoxinfang@cuhk.edu.hk (Guoxin Fang).}

% \email{}
\affiliation{%
  \institution{The Chinese University of Hong Kong / Centre for Perceptual and  Interactive Intelligence}
  \city{Hong Kong}
  \country{China}
}

% \author{Guoxin Fang}
% \email{guoxinfang@cuhk.edu.hk}
% \affiliation{%
%   \institution{The Chinese University of Hong Kong}
%   \city{Hong Kong}
%   % \state{Ohio}
%   \country{China}
% }

%\renewcommand{\shortauthors}{Qu et al.}

\begin{abstract}
We introduce a general, scalable computational framework for multi-axis 3D printing based on implicit neural fields (INFs) that unifies all stages of toolpath generation and global collision-free motion planning. 
In our pipeline, input models are represented as signed distance fields, with fabrication objectives—such as support-free printing, surface finish quality, and extrusion control—directly encoded in the optimization of an implicit guidance field. This unified approach enables toolpath optimization across both surface and interior domains, allowing shell and infill paths to be generated via implicit field interpolation.
The printing sequence and multi-axis motion are then jointly optimized over a continuous quaternion field. Our continuous formulation constructs the evolving printing object as a time-varying SDF, supporting differentiable global collision handling throughout INF-based motion planning.
Compared to explicit-representation-based methods, INF-3DP achieves up to two orders of magnitude speedup and significantly reduces waypoint-to-surface error. We validate our framework on diverse, complex models and demonstrate its efficiency with physical fabrication experiments using a robot-assisted multi-axis system.
%GPU-parallel querying directly assigns optimized waypoints, sequences, and motion frames for robot-assisted fabrication. 
%showing that the computational results enable support-free fabrication, high-quality surface finishing, and a globally collision-free printing process.
\end{abstract}




% \received{20 February 2025}
% \received[revised]{12 March 2025}
% \received[accepted]{5 June 2025}


\begin{teaserfigure}
\centering
    \includegraphics[width=1.0\textwidth]{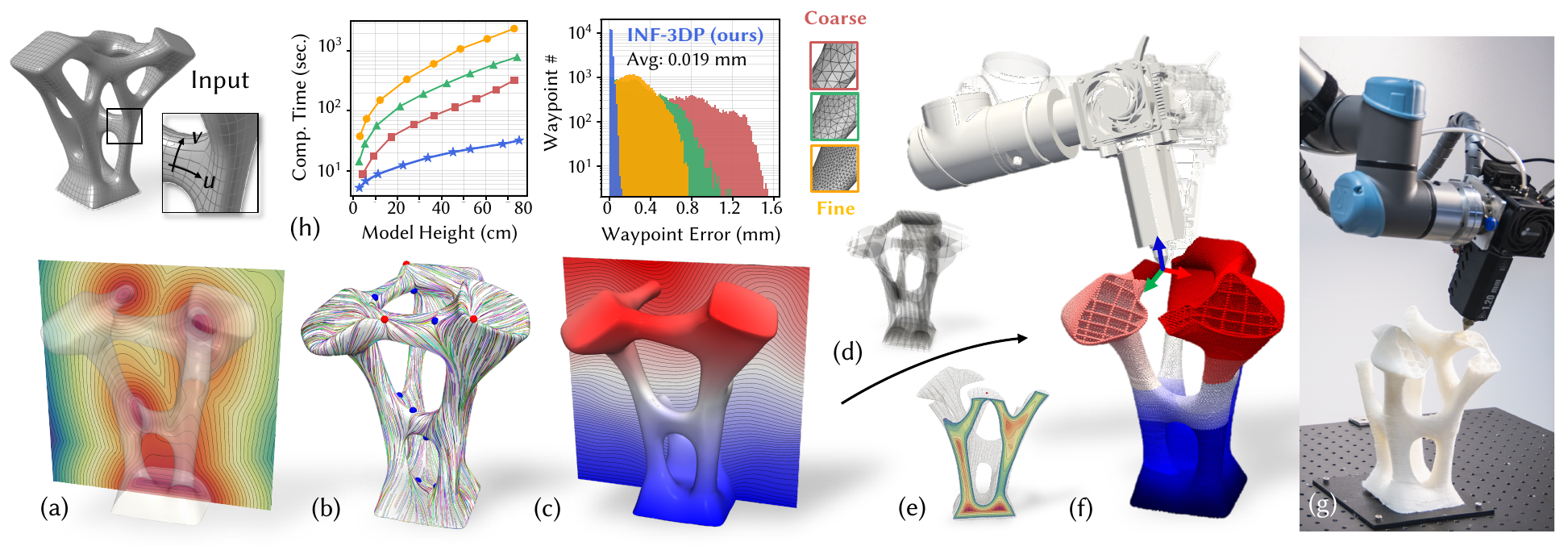}
    \caption{
We present INF-3DP, a computational framework leveraging Implicit Neural Fields (INFs) to optimize toolpath generation and plan global collision-free motion for multi-axis 3D printing. (a) \rev{Given a parameterized surface model as input, }{Given various types of geometric model as input}, signed distance field (SDF) is used as a volumetric representation, which enables (b, c) seamless optimization of a guidance field across both 2-manifold (surface) and 3-manifold (volume) domains with fabrication-aware objectives. Toolpaths for both the shell and infill structures are then generated using field interpolation based on the guidance and (d) infill fields. (e) Effective modeling of the dynamically evolving printed object via process-dependent time-varying SDF, which supports (f) differentiable global collision-free motion planning. (g) The computational result drives successful physical fabrication with a robot-assisted setup. 
(h) %With GPU-parallel processes in toolpath generation and motion planning with INF representation, 
Compared to existing resolution-dependent mesh-based solutions, our method offers superior computational scalability for large-scale problems while maintaining high precision in the generated waypoints.
    }
    % \Description{figure description}
    \label{teaser}
\end{teaserfigure}

\maketitle

\input{tex/intro}

\input{tex/relatedWork}
\input{tex/overview}

\input{tex/printField}

\input{tex/implementation}

\input{tex/result}
\input{tex/conclusion}

\begin{acks}
The authors would like to thank the anonymous reviewers for their valuable comments. This research was supported by the CUHK Institute of Intelligent Design and Manufacturing, CUHK Direct Research Grant (CUHK/4055223), the Shun Hing Institute of Advanced Engineering at CUHK (MMTp12-25), the HKSAR Research Grants Council Early Career Scheme (RGC-ECS) (CUHK/24204924). This research is also supported in part by the Centre for Perceptual and Interactive Intelligence, a CUHK-led InnoCentre under the InnoHK initiative of the Innovation and Technology Commission of the Hong Kong Special Administrative Region Government.
\end{acks}

\bibliographystyle{ACM-Reference-Format}
\bibliography{sample-base}

%\appendix

\end{document}

%% file: tex/intro.tex
\section{Introduction}
\label{sec:intro}
%\Guoxin{--- Rewrite on May 4th ---}
%\Guoxin{--- Go back again on May 17th ---}

% brief intro of multi-axis 3DP
Multi-axis 3D printing (3DP) has gained increasing attention for overcoming the limitations of conventional planar-based solutions with fixed nozzle direction. By dynamically adjusting the local printing direction (LPD) with high degrees-of-freedom (DOFs) setups~\cite{urhal2019robot}, multi-axis 3DP enables advanced functionalities such as support-free printing~\cite{dai2018support}, enhanced surface quality~\cite{etienne2019curvislicer}, and anisotropic property control~\cite{fang2020reinforced}. 
%Meanwhile, the usage of robot arms and other multi-axis platforms expand workspace and allow fabrication of complex geometries beyond the reach of conventional planar-based solutions~\cite{urhal2019robot}. 
%\subsection{Challenges of Computation for Multi-Axis 3DP}
% Computational task

Although high-DOF systems offer greater flexibility for material accumulation, they substantially increase the computational complexity required for successful multi-axis printing.
Algorithms must decompose the model into spatially defined toolpaths that satisfy fabrication-aware objectives, and subsequently plan multi-axis motions to ensure collision-free printing. For toolpath generation, field-based methods have shown promise in modeling material accumulation sequence in space, allowing various objectives to be encoded for optimization~\cite{zhang2022s3, li2022vector}. On the other hand, effective collision-free motion planning for multi-axis systems remains an open problem. Since the shape of the fabricated object changes dynamically throughout the printing process, the global collision checking between the object and the multi-axis system can be computationally intensive~\cite{dai2020planning}. This complexity is further compounded by the large number of waypoints—often in the millions—required for multi-axis printing, making it highly challenging to achieve robust and scalable planning with a discrete point-based solution~\cite{chen2025co}.

Existing approaches for spatial toolpath planning primarily rely on \textit{explicit} representations (e.g., voxels~\cite{dai2018support}, surface meshes~\cite{mitropoulou2020print}, tetrahedra~\cite{fang2020reinforced}, and hexagon meshes~\cite{cam2023fluid}) to discretize the computational domain. However, such tessellation introduces several computational challenges for multi-axis 3DP, including:

\begin{enumerate}
\item \textit{Hard to balance accuracy and efficiency:} explicit representations are resolution-dependent. With printing size scale-up, maintaining waypoint precision requires more elements, which dramatically increases computation time (see Fig.~\ref{teaser}(h)). 

\item \textit{Challenges of optimization across domains:} Existing methods tend to use surface meshes with objectives defined on \textit{2-manifolds} for shell structures~\cite{mitropoulou2024fabrication,zhong2023continuous}, or volumetric meshes with objectives defined on \textit{3-manifolds} for solid structures~\cite{li2021multi}. Domain-specific representations make these solutions hard to generalize across cases (details presented in Sec.~\ref{sec:objective}).

\item \textit{Difficulties in global collision handling:} 
The shape of the printed object changes dynamically during the printing sequence, making it difficult for explicit representations to efficiently capture these variations. As a result, global collision checking becomes time-consuming and prevents effective optimization for subsequent motion planning.
\end{enumerate}
Due to these limitations, explicit representations constrain the generality and scalability of toolpath generation and motion planning for multi-axis 3DP. In this work, we present a general framework that leverages implicit neural field (INF) representation to overcome these computational barriers.

\subsection{Our Method and Contribution}
\label{subsec:contribute}
\begin{table*}[!t] 
    \color{black}
    \footnotesize
    \centering
    \caption{\textcolor{black}{Comparison of related work on computational pipeline for robot-assisted multi-axis 3DP.} %\Aoran{"Representation" is more precise than "Comp. Domain" here.}
    } 
    \label{table:comparsion}
    %\vspace{-5px}
    \begin{tabular}{l|l||c|c|c|c|c}
        \hline
        \multirow{2}{*}{{Methods}} & \multirow{2}{*}{{Representation}} & \multicolumn{5}{c}{{Capabilities}} \\ \cline{3-7}
        & & {Printing Type}  & {Local Gouging} & {Global Collision-Free} & {Parallel Comp.} & {Waypoint Error} \\ \hline
        \cite{dai2018support} & Voxel & \cellcolor{green!12} Shell + Solid  & \cellcolor{red!12} No & \cellcolor{green!12} Yes & \cellcolor{red!12} No & \cellcolor{red!12} High \\ 
        \cite{mitropoulou2020print} & Surf. Mesh & \cellcolor{red!12} Shell  & \cellcolor{red!12} No & \cellcolor{red!12} No & \cellcolor{red!12} No & \cellcolor{gray!10} Res. Dependent \\
        \cite{zhang2022s3} & Vol. Mesh & \cellcolor{red!12} Solid  & \cellcolor{green!12} Yes & \cellcolor{red!12} No & \cellcolor{red!12} No & \cellcolor{gray!10} Res. Dependent \\ 
        \cite{liu2024neural} & INF + Vol. Mesh & \cellcolor{green!12} Shell + Solid  & \cellcolor{green!12} Yes & \cellcolor{red!12} No & \cellcolor{red!12} No & \cellcolor{gray!12} Res. Dependent \\ 
        INF-3DP (Ours) & INFs & \cellcolor{green!12} Shell + Solid  & \cellcolor{green!12} Yes & \cellcolor{green!12} Yes & \cellcolor{green!12} Yes (GPU-based) & \cellcolor{green!12} Low (Res. Independent) \\ 
        \hline
    \end{tabular} 
\end{table*}
%\Guoxin{The advantage of better field optimization formulation needs to be added here.}
As illustrated in Fig.~\ref{teaser}, several key INFs that encode geometry, material deposition sequence, and motion planning objectives are optimized throughout the pipeline.
A signed distance field ($\phi: \mathbb{R}^3 \rightarrow \mathbb{R}$, Fig.~\ref{teaser}(a)) is first used as a continuous volumetric representation for the input model $\mathcal{M}$, then a guidance field ($g: \mathbb{R}^3 \rightarrow \mathbb{R}$, Fig.~\ref{teaser}(b,~c)) is optimized by encoding objectives in both model surface $\mathcal{S}$ and interior region $\Omega$. 
%($\mathcal{S} \subseteq \mathbb{R}^3 := \{\mathbf{x}  \mid \phi(\mathbf{x}) = 0\}$ and  $\Omega \subseteq \mathbb{R}^3 := \{\mathbf{x}  \mid \phi(\mathbf{x}) < 0\}$). 
Together with a set of optimized infill fields ($\mathcal{\psi}: \mathbb{R}^3 \rightarrow \mathbb{R}$, Fig.~\ref{teaser}(d)), toolpath for shell and solid structure can be extracted implicitly by field interpolation (detail presented in Sec.~\ref{sec_toolpath}).
%(w.r.t., iso-values $u, v$) for shell and solid structure is define implicitly as $\mathcal L_{u}^\mathcal{S} = ~\{\mathbf{x} \mid g(\mathbf{x}) = u,\phi(\mathbf{x}) = 0 \}$ and $\mathcal L_{uv}^{\Omega} = ~\{\mathbf{x} \mid g(\mathbf{x}) = u, \psi(\mathbf{x}) = v,\phi(\mathbf{x}) < 0 \}$, respectively. 
In the stage of motion planning, a sequence field ($\mathcal{T}: \mathbb{R}^3 \rightarrow \mathbb{R}$) is used to encode printing order, where the geometry of the dynamically growing printing object $\mathcal{M}_{\mathcal{T}}$ is presented by time-varying SDF ($\phi_{\mathcal{T}}$, Fig.~\ref{teaser}(e)). The printing sequence is jointly optimized with the motion of the multi-axis system, represented by a quaternion field ($\mathbf{q}: \mathbb{R}^3 \rightarrow \mathbb{R}^4$), to ensure an as-continuous-as-possible smooth printing motion while maintaining global collision avoidance (Fig.~\ref{teaser}(f)). %his enables successful fabrication by a robot-assisted setup as shown in Fig.~\ref{teaser}(g).

Inviting implicit neural fields into toolpath generation and motion planning offers significant advantages.
Firstly, this continuous representation facilitates the integration of \rev{}{fabrication-aware objectives across computational domains.
%\Aoran{1) both volumetric and co-dimensional domain? 2) both full-dimensional and co-dimensioanl domain? 3) both 2-manifold and 3-manifold domain? 1 and 2 are a little bit better from my view but introduce new terminologies.
Objectives such as support-free and high-quality surface finishing (defined in $\mathcal{S}$), as well as singularity-free for manufacturability (defined in $\Omega$) are jointly optimized using a two-stage learning scheme (detailed in Sec.~\ref {subsec:guidance}). This provides a unified solution to compute optimized spatial toolpath for both shell and solid models with controllable infill structure.} 
On the other hand, the INF representation overcomes the computational bottlenecks of global collision checking for dynamically evolving printed shapes through the evaluation of time-varying SDF (Fig.\ref{teaser}(e), details presented in Sec.~\ref{subsec:Time-SDF}), and support differentiable motion planning by continuous optimization of the quaternion field. 
With implicitly-defined toolpaths through field intersections, our method generates printing waypoints $\mathcal{P}$ that accurately capture fine geometric details via efficient, highly parallel discretization with field projection (see Sec.~\ref{sec_toolpath}). The printing sequence $\mathcal{T}$ and motion $\mathbf{q}$ for these waypoints can be directly determined through efficient INF querying, ensuring excellent computational scalability and enabling high-precision 3DP with complex geometries.

Unlike most recent work of Neural-Slicer~\cite{liu2024neural}, which optimizes implicitly-defined guidance fields but still relies heavily on tetrahedral meshes for toolpath generation, our approach uses implicit neural fields as a unified representation for printed models, guidance fields, and machine motion in all computational stages for multi-axis 3DP. This eliminates resolution dependence, enables differentiable optimization, and overcomes the scalability challenge. \rev{}{A comparison with existing works is summarized in Table~\ref{table:comparsion}, and} as shown in Fig.\ref{teaser}(h), INF-3DP achieves up to two orders of magnitude speedup and significantly lower waypoint-to-surface error compared to mesh-based solutions. The technical contributions of this work are summarized as follows: 

\begin{itemize}
    \item Introduced a scalable computational framework based on neural implicit field representation, providing a general 
    %toolpath generation and global collision-free motion planning 
    solution for filament-based multi-axis 3DP.
    
    %\item Systematically formulated objectives for multi-axis 3DP and proposed a two-stage guidance field computing that seamlessly optimizes objectives defined on both \textit{2-manifold} and volumetric (\textit{3-manifold}) domains.
    \item Proposed a two-stage guidance field optimization method that seamlessly \rev{}{integrates and optimizes} fabrication-aware objectives on \textit{2-manifold} and \textit{3-manifold}, generates \rev{optimal toolpath}{spatial toolpath for shell and infill} while ensuring manufacturability.

    \item Take advantage of implicitly defined toolpaths, conduct highly parallel discretization with field projection to efficiently generate high-precision printing waypoints. 

    \item Proposed an implicit formulation for the dynamically evolving printed object as time-varying SDF by field interpolation, enabling efficient collision evaluation and differentiable motion planning through quaternion field optimization.
    
    %\Guoxin{To discuss.} \Js{resolving intersections of implicit fields, not only guidance fields.}
    
\end{itemize}
To the best of our knowledge, this is the first computational pipeline to systematically address toolpath generation for both shell and solid structures while enabling scalable global collision avoidance in multi-axis 3DP motion planning. Leveraging GPU-parallel computing, our method offers a general and scalable solution for complex, large-scale cases. We validate our framework on various models, demonstrating clear advantages over mesh-based solutions. Physical experiments on a robot-assisted multi-axis 3DP system (see Fig.~\ref{teaser}(g)) further confirm its effectiveness.

%% file: tex/relatedWork.tex
\section{Related work}
In this section, we review the literature on toolpath generation and motion planning for multi-axis 3D printing, and the recent advances in the use of implicit neural fields for graphics applications.

\subsection{Multi-axis 3DP: Application and Toolpath Generation}

Multi-axis 3D printing is a rapidly growing research area in both the manufacturing and graphics communities, demanding advanced computational methods to fully leverage multi-axis motion and address the limitations of conventional planar-based approaches, such as weak mechanical strength \cite{ahn2002anisotropic}, poor surface quality \cite{chakraborty2008extruder}, and the need for support structures \cite{zhang2015perceptual}. Prior works have shown that with a designed spatial toolpath, multi-axis 3DP can eliminate support structures~\cite{huang2016framefab, dai2018support, mitropoulou2025non}, minimize the staircase effect for improved surface finishing~\cite{etienne2019curvislicer}, and enhance mechanical strength~\cite{tam2017additive, fang2024exceptional, liu2025neural}. 

When computing toolpaths with fabrication-aware objectives, existing methods typically rely on explicit representations, which are resolution-dependent and make it challenging to balance computational efficiency with geometric accuracy of fine model details~\cite{zou2014iso, rabinovich2017scalable}. Additionally, prior work treats shell and solid structures separately: for shell printing, recent studies emphasize field smoothness for high-quality surface finishing~\cite{shao2024dual, mitropoulou2025curvature} and toolpath continuity for manufacturability~\cite{hergel2019extrusion, lau2023partition, zhong2023continuous}, while solid printing objectives are typically defined per volumetric element~\cite{fang2020reinforced, zhang2022s3}, with both layer-based~\cite{liu2024neural, xu2019curved, zhang2022s3} and layer-free~\cite{cam2023fluid} approaches reported. To date, a unified and scalable framework for both shell and solid multi-axis 3DP remains lacking. We address this with the help of implicit neural field representation~\cite{sitzmann2020implicit}, providing a computational pipeline that enables toolpath generation for support-free multi-axis 3DP with high-quality surface finishing.

\subsection{Collision-Free Planning for Multi-Axis Printing}

Motion planning with collision-free constraints remains a major challenge in multi-axis manufacturing~\cite{tang2024review}. As the workpiece changes shape during fabrication, collision checking must be performed dynamically throughout the process~\cite{dai2020planning, dutta2024vector}, which is computationally intensive. Early approaches, such as convex-hull-based voxel growing~\cite{dai2018support}, encode collision constraints but greatly restrict feasible motions. For multi-axis 3D printing of truss structures, sampling-based detection and graph search are often used to plan collision-free paths~\cite{huang2016framefab, wang2023temporal}. Mesh-based methods are also considered, which typically model the nozzle as a convex hull and detect collisions by checking for mesh intersections with already printed parts~\cite{fang2020reinforced, zhang2021singularity}. In subtractive manufacturing, partition-based strategies are invited to help reduce global collision risks~\cite{zhao2018dscarver, zhong2023vasco}.

Recent advances have introduced data-driven techniques for accessibility learning~\cite{dai2020planning, zhong2025deepmill}. Reinforcement learning (RL) has also been applied to motion planning for multi-axis 3D printing~\cite{huang2024learning} and robot-assisted assembly~\cite{wang2025learning}. In the broader robotics field, implicit-SDF has been used to approximate robot arm geometry~\cite{li2024representing} and enable fast collision checking with motion planning~\cite{chen2022fully, koptev2022neural}. In this work, we leverage INF representations for efficient collision checking in multi-axis 3D printing. By checking the sign and value of the time-varying SDF (see Sec.~\ref{subsec:SDF}), we can rapidly detect collisions and compute collision loss, enabling differentiable and continuous motion planning through INF-based optimization.

%Differently, unlike existing work generally train an end-to-end network for time-varying SDF, we conduct xxx precise xxx (detail discussed in Sec.~\ref{}). 

\subsection{Field-based Optimization and Implicit Neural Field}

Field-based optimization is widely used for graphics tasks such as high-quality mesh generation~\cite{ni2018field, liu2024neural}, deformation control~\cite{von2006vector, liu2021knitting}, computational design~\cite{arora2019volumetric, montes2023differentiable}, \rev{}{and digital fabrication~\cite{mitra2023helix, ma2021stylized, ma2020designing}}. 
As a general practice, fields are encoded on discrete domains—such as triangle meshes for surfaces~\cite{azencot2015discrete} or tetrahedral meshes for volumes~\cite{zhu2025designing}—with objectives like field smoothness~\cite{knoppel2013globally, de2016vector} or alignment with boundary conditions~\cite{huang2011boundary} being optimized using discrete differential geometry (DDG) methods. In this work, the problem of toolpath and motion planning for multi-axis 3DP is formulated as a field optimization problem (presented in the next section); differently, we adopt a learning-based approach to optimize implicit fields that enables seamless integration of objectives defined in surface and volumetric domains.

%For field optimization with optimization in both domains, surface objectives are generally considered as boundary conditions rather than optimized together in a 3-manifold.

Recent advances in implicit neural fields have brought significant breakthroughs in visual computing~\cite{xie2022neural}, enabling implicit encoding of geometry as a signed distance function~\cite{park2019deepsdf, wang2023neural, wang2024implicit}. INF have shown promise for solving PDEs~\cite{sitzmann2020implicit, raissi2019physics}, mesh generation~\cite{dong2024neurcross, chang2024neural}, and design optimization~\cite{zhang2025dualms}. They also benefit toolpath planning for both planar~\cite{wade2025implicit} and multi-axis 3D printing~\cite{liu2024neural}, and support neural-based collision checking~\cite{zhong2025deepmill}. In our pipeline, INFs serve as unified representations for models, guidance fields, and machine motion at all computational stages. This allows end-to-end differentiable optimization and removes the scalability and resolution constraints inherent to mesh-based methods.

%% file: tex/overview.tex
\section{Overview and Objectives for Field Computing}
\label{sec:objective}
%\Guoxin{Working on this first --- May 2rd}
%\Guoxin{Finalized on May 6nd --- }

\begin{figure}[!t]
    \centering
    \includegraphics[width=1.0\linewidth]{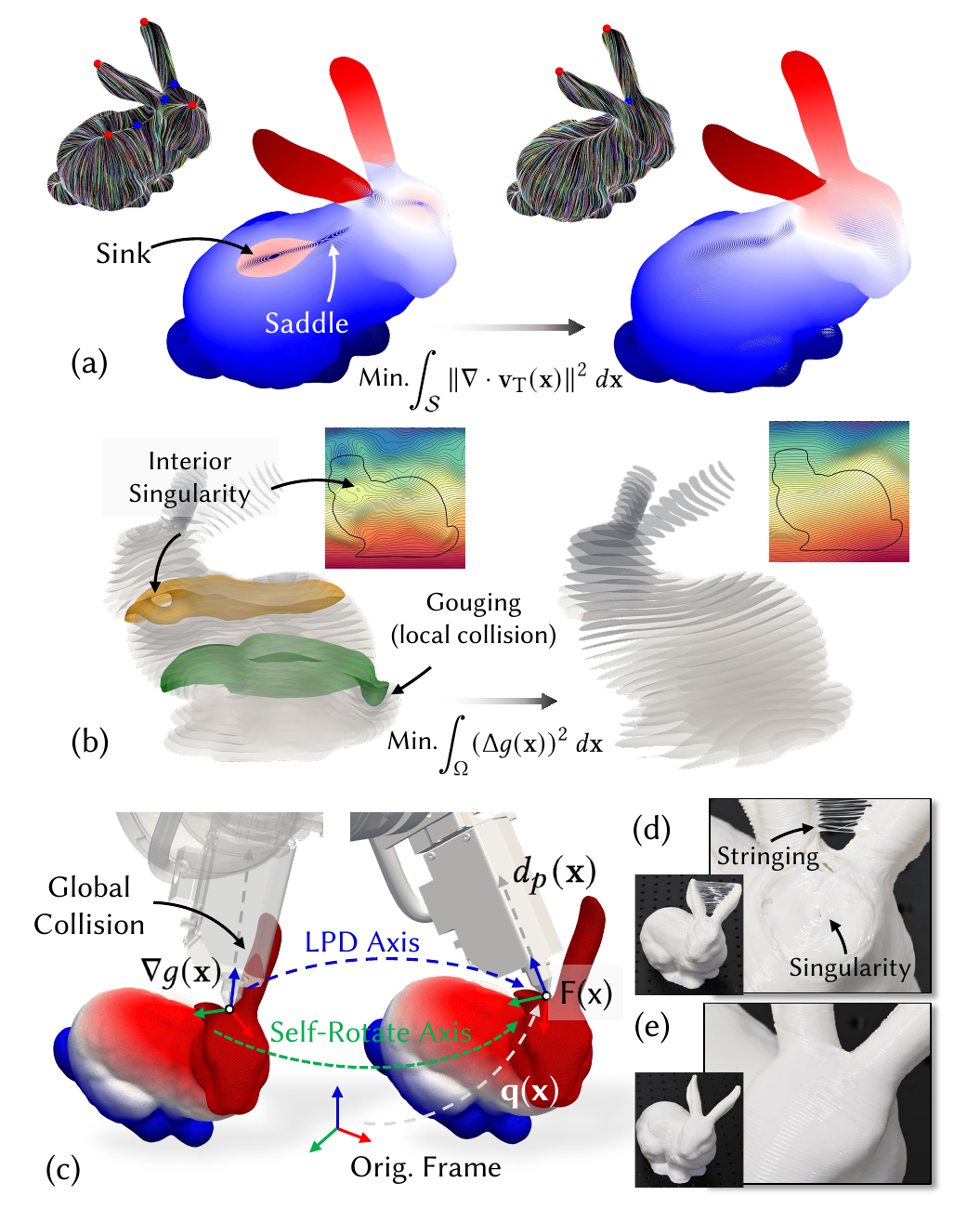}
    \caption{
    % Fabrication-related objectives in multi-axis 3D printing and their relationship to field optimization:
    % (a) The placement of singularities in the vector field $\mathbf{v}_{\text{T}}$ on the \textit{2-manifold} $\mathcal{S}$ is crucial for printing quality of the surface.
    % (b) Similarly, manufacturability of the infill structure in the \textit{3-manifold} $\Omega$ requires the guidance field $g(\mathbf{x})$ to be smooth and free of singularities to prevent local collisions. 
    % (c) Left: Directly applying $\nabla g(\mathbf{x})$ as the LPD can cause global collision; Right: collision-free planning is achieved by determining a feasible frame field $\mathbf{F}(\mathbf{x})$ (encoded by the quaternion vector $\mathbf{q}(\mathbf{x})$), which explicitly incorporates both the LPD axis and the self-rotation axis.
    % (d) Illustration of the stringing issue with discontinuous motion, and significant fabrication error caused by singularity. 
    % (e) By considering all fabrication-related objectives, we achieve support-free printing for the bunny and obtain superior surface finishing.
Fabrication objectives for multi-axis 3DP are closely related to field optimization:
(a) Singularity placement of the vector field $\mathbf{v}_{\text{T}}$ is essential for surface quality.
(b) Similarly, optimize for a smooth, singularity-free guidance field $g(\mathbf{x})$ ensure manufacturability and prevent collisions.
(c) Directly applying $\nabla g(\mathbf{x})$ as the LPD can lead to global collisions, whereas optimizing a feasible frame field $\mathbf{F}(\mathbf{x})$—which accounts for both the LPD and self-rotation axes through quaternion $\mathbf{q}(\mathbf{x})$—enables collision-free planning.
(d) Discontinuous motion and singularities cause stringing and significant fabrication errors.
(e) By integrating all objectives, we achieve support-free, high-quality printing of the bunny model.}
\label{fig_objective}
\end{figure}

This section first presents preliminary concepts of computation, highlighting how fabrication-related objectives (illustrated in Fig.~\ref{fig_objective}) for both shell and solid multi-axis 3DP can be formulated as optimization problems for INFs. We then provide a short discussion on field optimization as the key to the proposed computational pipeline.

\begin{figure*}[t]
    \centering
    \includegraphics[width=1.0\linewidth]{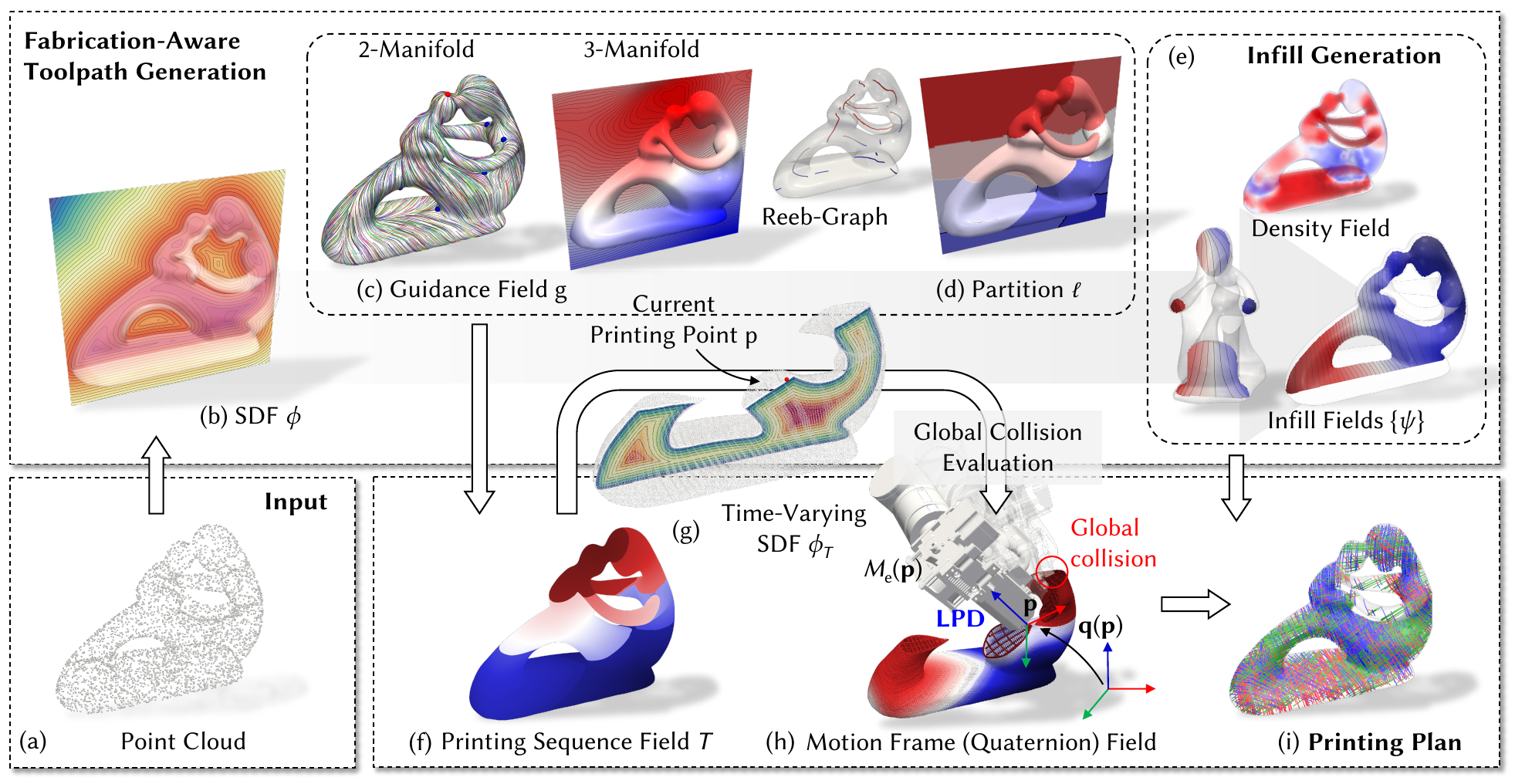}
    \caption{The computational pipeline of INF-3DP. (a) Begins with representing the printing model $\mathcal{M}$ as a point cloud for training, (b) a SDF $\phi$ is computed as an implicit volumetric representation of the model. (c) Guidance fields are then optimized with fabrication-aware objectives on both the surface (2-manifold) and interior (3-manifold) domains, (d) guiding Reeb-graph-based partitions (indexed by $\ell$) to ensure printing continuity. (e) Infill and density fields are constructed for variable density infill, (f) and the optimized printing sequence field $\mathcal{T}$ is established based on $g$ and $\ell$. (g) By evaluating the time-varying SDF $\phi_{\mathcal{T}}(\mathbf{x})$, the shape of the object at each stage of printing is approximate, enabling efficient global collision detection. (h) Differentiable quaternion field optimization is then applied for collision-free and smooth motion planning. (I) The final fabrication plan, visualized as streamlines of the motion frame field, integrates all optimized fields to ensure support-free, high-quality, and collision-free multi-axis 3DP.}
    \label{fig_pipeline}
\end{figure*}

As previously mentioned in Sec.~\ref{subsec:contribute} and illustrated in Fig.~\ref{fig_pipeline}, multiple INFs are optimized through learning by considering fabrication-related objectives, which include \rev{}{
\begin{itemize}
    \item SDF $\phi$ and time-varying SDF $\phi_{\mathcal{T}}(\mathbf{p})$ represent the entire object and its growth at waypoints $\mathbf{p}$ during printing;
    \item the guidance field $g$ and infill fields $\psi$ for toolpath generation;
    \item printing sequence $\mathcal{T}$ with quaternions $\mathbf{q}$ for motion planning.
\end{itemize}
In our framework, SDF implicitly defines the continuous computational domain of surface $\mathcal{S} \subseteq \mathbb{R}^3 := \{\mathbf{x}  \mid \phi(\mathbf{x}) = 0\}$ and interior region $\Omega \subseteq \mathbb{R}^3 := \{\mathbf{x}  \mid \phi(\mathbf{x}) < 0\}$).}
%of surface $\mathcal{S} = \partial \mathcal{M}$ and interior region $\Omega$. 
For a point $\mathbf{x}\in\mathbb{R}^3$, $\nabla g(\mathbf{x})$ controls the growing direction of infill structure, while the material accumulation on the surface $\mathcal{S}$ is encoded by the tangent vector $\mathbf{v}_{\mathrm{T}}(\textbf{x})$. This vector is obtained by projected from $\nabla g(\textbf{x})\in \mathbb{R}^3$ onto the local tangent plane $T_{\mathbf{x}}$ on \textit{2-manifold} by projection operator $f_{\text{proj}}(\cdot)$ as
\begin{equation}
\label{eq_proj}
\mathbf{v}_{\mathrm{T}}(\textbf{x}) =  f_{\text{proj}}(\nabla g(\textbf{x})) = \nabla g(\textbf{x}) - \frac{\nabla g(\textbf{x}) \cdot \nabla \phi(\textbf{x})}{\|\nabla \phi(\textbf{x})\|^2} \nabla \phi(\textbf{x}).
\end{equation}
As the vector $\mathbf{v}_\text{T}$ can be analytically expressed from $g$, objectives for toolpath generation discussed below for surface $\mathcal{S}$ and interior domain $\Omega$ can be seamlessly integrated in the guidance INF. 
%This is challenging for explicit representations-based solutions as cross manifold objectives ....

%Unlike prior work focusing solely optimize toolpath and motion for shell models~\cite{zhong2023continuous} or solid models~\cite{zhang2022s3, liu2024neural}, our pipeline provides a unified solution for both cases, and the comparison with shell/solid bunny model is presented in Fig.~\ref{fig_objective}. In our pipeline, $\nabla g(\mathbf{x})$ controls printing sequence inside the model (i.e., the growing direction of infill structure), and the material accumulation direction on the surface $\mathcal{S}$ is represented by tangent vector $\mathbf{v}_{\mathrm{T}}(\textbf{x})$ that is projected from $\nabla g(\textbf{x})\in \mathbb{R}^3$ to the local tangent plane $T_{\mathbf{x}}$ on \textit{2-manifold} by projection operator $\mathcal{T}_{\text{proj}}(\cdot)$ as
% \begin{equation}
% \label{eq_proj}
% \mathbf{v}_{\mathrm{T}}(\textbf{x}) =  \mathcal{T}_{\text{proj}}(\nabla g(\textbf{x})) = \nabla g(\textbf{x}) - \frac{\nabla g(\textbf{x}) \cdot \nabla \phi(\textbf{x})}{\|\nabla \phi(\textbf{x})\|^2} \nabla \phi(\textbf{x}).
% \end{equation}
%With the help of INFs representation in both model and guidance field, we can continuously elaborate the objectives in surface $\mathcal{S}$ and interior domain $\Phi$ with the help of $\mathcal{N}_{\text{SDF}}$ and $\mathcal{N}_{\text{guide}}$, which is challenging for explicit-based solutions. \Guoxin{This need to discuss with Aoran.}

\subsection{Multi-Axis 3DP: Objectives for Field Computing}

In this work, fabrication-oriented objectives—including support-free printing, surface quality improvement, and local thickness control—are considered for shell and infill toolpath generation. Additionally, both local and global collision avoidance are ensured, while motion continuity and smoothness are optimized.
%All these objectives and constraints in multi-axis 3DP are tightly integrated into the field computations of our proposed pipeline.

\setlength{\intextsep}{0pt} % Adjusts vertical spacing
\setlength{\columnsep}{10pt} % Adjusts horizontal spacing
\begin{wrapfigure}{r}{0.4\linewidth} % 'r' for right, 'l' for left
\includegraphics[width=\linewidth]{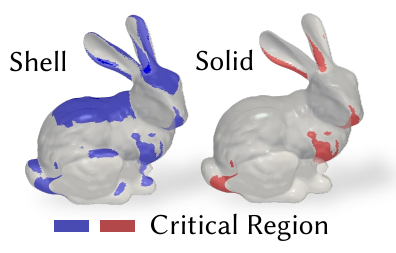}
\end{wrapfigure}
\subsubsection{Support-Free Printing} \label{subsubsection:support}
For both shell and infill printing, the support-free condition~\cite{zhang2022s3, liu2024neural} is related to the angle between the local printing direction and the surface normals. As a general pipeline, we consider a stronger condition for shell structure, where both the inner and outer regions need to be printed in a support-free manner (as shown in the warp figure—see the back and ear parts of the bunny). The constraint is defined as:
\begin{equation}
\label{eq_sf}
\mathcal{C}_{\text{SF}}:=|d_{p}(\mathbf{x}) \cdot \mathbf{n}| - \sin(\alpha) \leq 0
\end{equation}
Here, boundary normal is computed with SDF $\textbf{n} = \nabla \phi (\mathbf{x}) / \|\nabla \phi (\mathbf{x})\|$ for surface and $\textbf{n} = \nabla \psi (\mathbf{x}) / \|\nabla \psi (\mathbf{x})\|$ for infill. $\alpha$ represents the support-free angle, which depends on the material properties and taken as $40^\circ$ for the~\textit{polylactic acid} (PLA) materials. Note that during the toolpath generation stage, $\nabla g$ is used to represent the LPD. This constraint must also be considered during the motion planning process, as the LPD will be optimized together with the quaternion motion field $\mathbf{q}$.
%Note that the local printing direction $d_p(\mathbf{x})$ does not necessarily have to align with $\nabla g(\mathbf{x})$, as this could cause global collisions during printing (see Fig.~\ref{fig_objective}(d)). Thus, $\mathcal{C}_{\text{support}}$ must be guaranteed alongside collision-free constraints when planning the motion for 3DP (i.e., optimization for $\mathcal{T}$ and $\mathbf{q}$).
%It is worth mentioning that the support-free objective is not influenced by the gravity direction~\cite{}, as demonstrated by an extreme case in Fig.~\ref{} involving a sculpture model. In this example, the printing direction is opposite to the z-axis, yet a high-quality, support-free print is still achievable using a bottom-up local printing direction.

\subsubsection{Printing Quality and Manufacturability by Field Smoothness}
%\Guoxin{remember to add the definition of singularity everywhere in the paper.} 
Conducting successful, high-quality multi-axis 3DP depends heavily on the smoothness and regularity of the guidance field.
For example, the location of singularities in the tangent vector field $\mathbf{v}_{\mathrm{T}}(\textbf{x})\in\mathcal{S}$ significantly affects the surface finishing. Here, a singularity is defined as a region where the vector field vanishes~\cite{nikolaev2013foliations}, and is detected by the condition $\|\mathbf{v}_{\mathrm{T}}(\mathbf{x})\| < \varepsilon$ ($\varepsilon$ is set to $10^{-2}$).  As shown in Fig.~\ref{fig_objective}(a), if a sink-type singularity~\cite{wang2023neural} occurs in critical region (visually or mechanically important - for example, the back of the bunny), it brings significant fabrication error (see Fig.~\ref{fig_objective}(d) and Fig.~\ref{fig:scan}(a)) and dramatically increases the likelihood of collisions due to the rapid directional changes in the printing. Conversely, a saddle-type singularity can also lead to severe under-accumulation region~\cite{zhang2021singularity}. For solid models with infill, singularities within the \textit{3-manifold} where $\nabla g$ vanishes can produce problematic regions. As shown in Fig.~\ref{fig_objective}(b), singularities lead to highly curved areas prone to local gouging and therefore not able to be printed~\cite {fang2020reinforced, dutta2024vector}.

%Therefore, computing a smooth vector field $\mathbf{v}_{\mathrm{T}}$ and a singularity-free guidance field is the key to optimizing the surface finishing quality and ensuring the manufacturability of shell and solid multi-axis 3DP. 
In this work, the placement of singularities on the surface and the smoothness within the interior domain are optimized by:
\begin{align}
     \mathcal{O}_{\text{smooth}}^\mathcal{S} (g) & 
     %= \frac{1}{2}~\int_{\mathcal{S}}  \| \nabla \mathbf{v}_{\mathrm{T}}(\mathbf{x}) \|^2 ~ d \mathbf{x} 
     = \frac{1}{2}~\int_{\mathcal{S}}  \| \nabla \cdot f_{\text{proj}}(\nabla g(\textbf{x})) \|^2 ~ d \mathbf{x}  \label{eq:smoothSurface}
     \\ \mathcal{O}_{\text{smooth}}^{\Omega} (g) & = 
\frac{1}{2}~\int_{\Omega} ( \Delta g(\mathbf{x}))^2 ~ d \mathbf{x},
%~\int_{\mathcal{S}} \| \nabla \cdot \mathbf{v}_{\text{T}}(\mathbf{x}) \|^2 ~ d \mathbf{x}. 
\label{eq:smoothInterior}
\end{align}
This approach also enables control over the local extrusion rate - a key factor in multi-axis 3DP - which is inversely proportional to the norm of the guidance field ($\| \mathbf{v}_{\mathrm{T}} (\mathbf{x}) \|$ on the surface and $\| \nabla g(\mathbf{x}) \|$ in the interior). Although for filament-based 3DP, extruder have the ability to dynamically adjust the extrusion ratio (typically between $[0.1R, 0.7R]$, where $R$ is the nozzle diameter)~\cite{etienne2019curvislicer, fang2020reinforced}, regions near singularities will easily exceed this range due to locally vanishing gradients. Optimizing the smoothness energies defined in Eq.~\ref{eq:smoothSurface} and Eq.~\ref{eq:smoothInterior} also regularizes the norm of the guidance field, therefore ensuring effective extrusion control.

%By optimizing the smoothness energies in Eq.~\ref{eq:smoothSurface} and Eq.~\ref{eq:smoothInterior}, we not only control the number and placement of singularities but also regularize the norm of the guidance field, therefore ensuring effective extrusion control (a result can be seen in Fig.~\ref{fig_statue}).

It is worth mentioning that the alignment between $\mathbf{v}_{\mathrm{T}}$ and $\nabla g$ needs to be guaranteed near the surface to maintain compatible printing directions of shell and infill. This is formulated as
\begin{equation}
    \mathcal{O}_{\text{align}} (g) = \int_\mathcal{S} \| \nabla g(\mathbf{x}) - f_{\text{proj}}(\nabla g(\textbf{x}))\|^2 d\mathbf{x} \label{eq:alignObjective}
\end{equation}
%\Aoran{maybe increase the weight near the surface is needed?}
In this way, the guidance field in the interior is smoothly connected to the surface tangent field, ensuring a consistent and feasible printing direction across the entire domain.

\subsubsection{Printing Continuity, Collision-Free, and Motion Smoothness}
\label{subsec:motionObjective}
In multi-axis 3DP, optimizing the printing sequence $\mathcal{T}$ and motion $\mathbf{q}$ is essential for achieving continuous, collision-free, and smooth operation. As illustrated in Fig.~\ref{fig_objective}(d), continuity in the printing motion significantly affects print quality~\cite{zhong2023continuous} - directly using $g$ as the printing sequence can cause large jumps between regions, leading to stringing issues~\cite{paraskevoudis2020real}. To mitigate this, we partition the printing toolpath into continuous regions using a Reeb-Graph, resulting in a minimal number of regions (region index denoted as $\ell$; see Fig.~\ref{fig_pipeline}(d)). Within each region, the printing sequence remains continuous following $g$, thereby reducing inter-region jumps. In this way, the printing sequence can be written as $\mathcal{T}(\mathbf{x}) = g(\mathbf{x}) + \ell(\mathbf{x})$, which is shown in Fig.~\ref{fig_pipeline}(f).

With dynamically changing printing direction, global collisions can happen between the already printed sections (represented by the time-varying SDF $\phi_{\mathcal{T}}$, see Fig.~\ref{fig_pipeline}(g)) and the printing setup. When the printing comes to waypoint $\mathbf{p}$, the geometry of the printing setup, including the extruder and connected robot joints, is denoted as $\mathcal{M}_{e}(\mathbf{p})$ \rev{}{and discretized as point clouds in our implementation}. A global collision can be detected if any point $\mathbf{x} \in \mathcal{M}_{e}(\mathbf{p})$ enters the printed model at a location where material has already been deposited, i.e., making the time-varying SDF $\phi_{\mathcal{T}}(\mathbf{p},\mathbf{x})) < 0$ (detail presented in Sec.~\ref{subsec:Time-SDF}). We formalize the global collision-free constraint for multi-axis 3DP as:
\begin{equation}
\mathcal{C}_{\text{coll}}(\mathbf{p}):= \int_{\mathcal{M}_{e}(\mathbf{p})}
H(-\phi_{\mathcal{T}}(\mathbf{p},\mathbf{x}))
~d\mathbf{x} \equiv 0,
\end{equation}
where $H(\cdot)$ is the Heaviside step function. Additionally, motion smoothness is also important to ensure compatibility of local printing direction between neighboring waypoints, which is optimized via the objective:
\begin{equation}
\mathcal{O}_{\text{smooth}}^{\mathcal{Q}} = \frac{1}{2} \int_{\mathcal{S}~\cup~\Omega} \| \nabla \mathbf{q}(\mathbf{x}) \|^2~d\mathbf{x}.
\end{equation}
As mentioned in Sec.~\ref{subsubsection:support}, the LPD in physical printing is adjusted from the initial $\nabla g$ according to the optimized $\mathbf{q}$. Therefore, the support-free constraints $\mathcal{C}_{\text{SF}}$ defined in Eq.~\ref{eq_sf} must also be incorporated into the motion planning process.

\subsection{Short Discussion on INFs Optimization} 
\label{subsec:shortDiscuss}

All fields in our framework are represented and optimized as INFs by learning with loss functions corresponding to the objectives described above. The choice of loss functions and training strategies strongly affects convergence and the quality of both toolpath generation and motion planning.

While the INF representation enables us to unify objectives defined on both the \textit{2-manifold} and \textit{3-manifold} domains into a single scalar field $g$, the smoothness, alignment, and support-free objectives can conflict with one another. Additionally, the optimization space for the guidance field is high-dimensional, and directly training with all objectives combined may lead to slow convergence and difficulty in balancing competing goals (see the ablation study in Sec.~\ref{subsec:guideAblation}). To address this, we adopt a two-step training procedure: first optimizing on the 2-manifold, then extending the field into the 3-manifold while enforcing the alignment objective on the boundary. The detailed training process is presented in Sec.~\ref{subsec:guidance}.

% For motion planning, evaluating the time-varying SDF is crucial for global collision checking. Rather than jointly training the SDF and printing sequence as a network—which can introduce errors in $\mathcal{C}_{\text{coll}}$ due to incorrect SDF signs—we use field interpolation between $\phi$ and $\mathcal{T}$ to compute $\phi_\mathcal{T}$, which guarantees correct sign evaluation to support strict collision-free planning. 
For motion planning, evaluating the time-varying SDF $\phi_\mathcal{T}$ is crucial for global collision checking. Rather than training an end-to-end time-dependent SDF network, which can introduce errors in $\mathcal{C}_{\text{coll}}$ due to incorrect SDF signs near the boundaries—we construct this by field interpolation between $\phi$ and $\mathcal{T}$, guaranteeing correct sign evaluation to support strict collision-free planning. 
The evaluation process is detailed with further discussion in Sec.~\ref{subsec:Time-SDF}. On the other hand, while graph-based partitioning optimizes for printing continuity, it can significantly increase the risk of global collisions, possibly leading to no feasible solution for a collision-free solution of $\mathbf{q}$. To address this, our pipeline iteratively refines the graph and updates the printing sequence $\mathcal{T}$, while optimizing motions $\mathbf{q}$ to strictly satisfy both $\mathcal{C}_{\text{coll}}$ and $\mathcal{C}_{\text{SF}}$. The detailed optimization process is presented in Sec.~\ref{subsec:motionTraining}.

%% file: tex/printField.tex
\section{Optimization for Implicit Neural Fields}

This section details the loss functions and optimization process for implicit neural fields. For the SDF, guidance field, infill fields, and quaternion field, the SIREN network structure~\cite{sitzmann2020implicit} is used as it demonstrated strong field representation capabilities. The specific network architectures and training procedures
%, and loss weightings 
are provided in Sec.~\ref{subsec:learningDetail}.

%\Js{For brevity, all hyperparameters of the loss terms are omitted in the loss functions and will be detailed in Sec.}

\subsection{Signed Distance Field}
\label{subsec:SDF}

%\Guoxin{A warp figure to illustrate the difference of ghost geometry loss need to be included - use bunny as example, also mention the point cloud input can be obtained }

Given a target model $\mathcal{M}$ represented in various geometric forms, the sampled point cloud is used as training data for SDF learning with loss defined as
%The SDF network $\mathcal{N}_{\text{SDF}}(\mathbf{x})$ is trained using the following loss terms 
\begin{equation}
    \begin{split}
    \mathcal{L}_{\mathrm{SDF}} = & \omega_{\text{dist}}
    \int_{\Psi_{0}} \| \phi(\mathbf{x}) \| d \mathbf{x}  + \omega_{\text{norm}} \int_{\Psi_{0}} \big( 1 - \langle \nabla \phi (\mathbf{x}) , \mathbf{n}(\mathbf{x}) \rangle\big) d \mathbf{x} \\ & + \omega_{\text{NM}} \int_{\Psi \backslash \Psi_0} \exp(-\gamma \cdot | \phi(\mathbf{x}) |) ~d \mathbf{x} \\ 
    &  +   \omega_{\text{eikonal}}\int_{\Psi} \Big \| \| \nabla \phi (\mathbf{x}) \| - 1 \Big \| d \mathbf{x}  + \omega_{\text{lap}} \int_{\Psi} \| \Delta \phi (\mathbf{x}) \| ^ 2 d \mathbf{x},
    \end{split}
\end{equation}
where the first two terms applied on the boundary $\Psi_0$
%near the model boundary $\Psi_0:= \{\mathbf{x}~|~|\phi(\mathbf{x}) - \epsilon | < 0\}$ \Aoran{I think the boundary definition should be point cloud-based}
, which enforce boundary distance and normal alignment respectively. 
%Here $\epsilon=0.01$ is used as the narrow band width. 
The third term is non-manifold loss, where $\gamma \gg 1$ is used to regularize regions far from the surface. The Eikonal and Laplacian losses are applied over the entire space $\Psi:= \{\mathbf{x}~|~\mathbf{x}\in\mathbb{R}^3\}$ to enforce unit gradient and minimize the divergence on $\nabla \phi(\mathbf{x})$~\cite{ben2022digs}. Notably, training the SDF without the final Laplacian term can already produce reliable results near the model surface, 
\setlength{\intextsep}{2pt} % Adjusts vertical spacing
\setlength{\columnsep}{10pt} % Adjusts horizontal spacing
\begin{wrapfigure}{r}{0.5\linewidth} % 'r' for right, 'l' for left
\includegraphics[width=\linewidth]{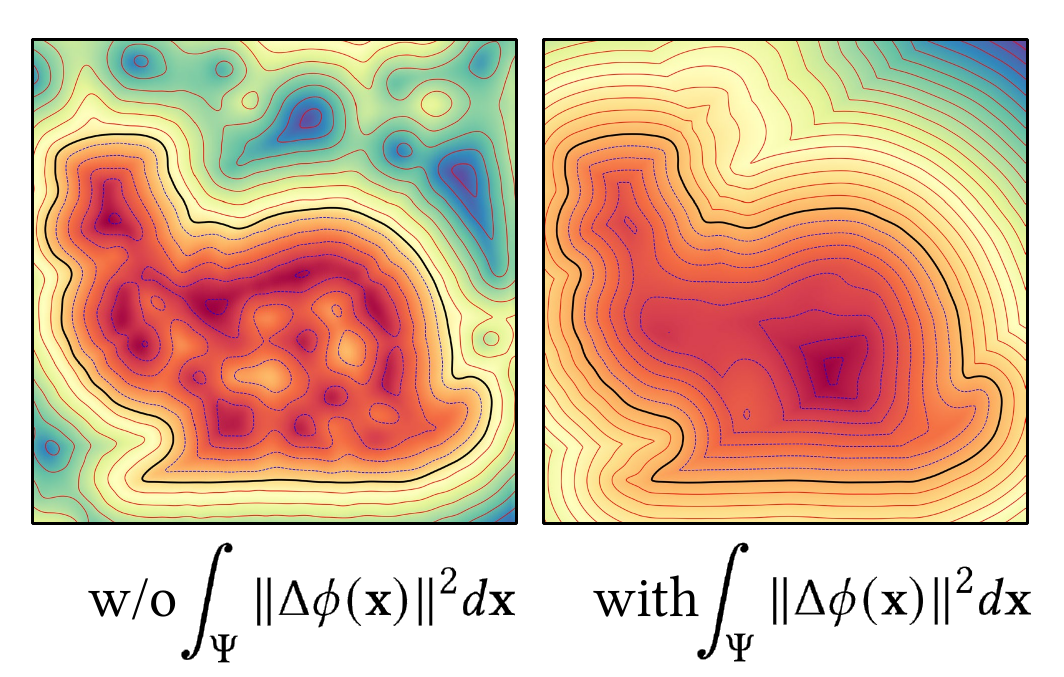}
\end{wrapfigure}
but it often leads to regions with local ghost geometry, as shown on the left of the warp figure. In our framework, the final Laplacian loss, although it increases training time due to the evaluation of $\Delta\phi$, is crucial for subsequent optimization. 
The implicit SDF needs to provide reliable analytical evaluation of geometric information for the interior space $\Omega$, which is essential for infill structure generation and for ensuring precise collision checking during motion planning.

\subsection{Guidance Field} 
\label{subsec:guidance}
%\Guoxin{This section should highlight two-step of the optimization - and explain why this is the case, and using one example here: maybe scanned scupture model is good! to highlight the difference between direct optimization and in-direction optimization - detail to discuss with Jiasheng.}

As discussed in Sec.~\ref{subsec:shortDiscuss}, the guidance scalar field that satisfies fabrication-related objectives is trained sequentially on the surface $\mathcal{S}$ and interior $\Omega$ to improve the convergence and make a good balance between objectives. For the surface domain, the loss used to optimize guidance INF is define as
\begin{equation}
        \mathcal{L}_\text{guide}^{\mathcal{S}}  = \omega_{\text{smooth}}^\mathcal{S} \mathcal{L}_{\text{smooth}}^{\mathcal{S}} + \omega_{\text{SF}}^\mathcal{S} \mathcal{L}_{\text{SF}}^\mathcal{S} + \omega_{\text{coll}}^{{\Omega}_{b}} \mathcal{L}_{\text{coll}}^{{\Omega}_{b}} \label{eq:guideSloss}
\end{equation}
where
\begin{align}
 \mathcal{L}_{\text{smooth}}^{\mathcal{S}} &=\int _{\mathcal{S}} \big \| \nabla 
    %\mathcal{T}_{\text{proj}}(\nabla g(\textbf{x}),\nabla \phi(\textbf{x}))
    \textbf{v}_{\text{T}}(\mathbf{x})\|^2 d\mathbf{x}, \\  \mathcal{L}_{\text{SF}}^\mathcal{S} &=   \int_{\mathcal{S}} \sigma \big( k_{\text{SF}} ( \big |\langle \frac{\nabla g(\mathbf{x})}{\|\nabla g(\mathbf{x}) \|} , \nabla \phi (\mathbf{x})  \rangle \big |  - \sin\alpha )    \big) d \mathbf{x}, \\ 
   \mathcal{L}_{\text{coll}}^{\Omega_b} &= \int _{\Omega_\text{b}} \big( 1 - \langle \frac{\nabla g(\mathbf{x})}{\|\nabla g(\mathbf{x})\|} , \mathbf{z} \rangle \big) d\mathbf{x}. 
\end{align}
\rev{}{The first smooth loss $\mathcal{L}_{\text{smooth}}^{\mathcal{S}}$ echoes the harmonic objective $\mathcal{O}_{\text{smooth}}^{\mathcal{S}}$ in Eq.~\ref{eq:smoothSurface},} where the vector $\textbf{v}_{\text{T}}(\mathbf{x})$ is computed by projecting gradient $\nabla g(\mathbf{x})$ onto the tangent plane of $\mathcal{S}$ by Eq.~\ref{eq_proj}. $\mathcal{L}_{\text{SF}}^\mathcal{S}$ is a relaxed term to encode support-free constraints \rev{}{$\mathcal{C}_{\text{SF}}$ (defined in Eq.~\ref{eq_sf}) into optimization}, where $\sigma(\cdot)$ is the sigmoid function and $k_{\text{SF}} = 30$ is used. $\mathcal{L}_{\text{coll}}^{\Omega_b}$ sets a Neumann boundary condition on the bottom region of the model $\Omega_b$ (controlled by height $h$) to enforce printing direction follow z-axis (i.e., $\mathbf{z} = [0,0,1]^\top$), which avoids local collision between model and printing platform. The Neumann boundary condition also avoids the training converting into a trivial solution (i.e., $g(\textbf{x}) \equiv 0$ in space). 

Subsequently, the loss for training guidance field in the interior domain $\Omega$ to further optimize guidance INF is defined as
\begin{equation}
    \mathcal{L}_\mathrm{guide}^{\Omega} = \omega_{\text{smooth}}^{\Omega} \mathcal{L}_{\text{smooth}}^{\Omega} + \omega_{\text{align}}\mathcal{L}_{\text{align}}  \label{eq:guidePhiloss}
\end{equation}
where
\begin{align}
\mathcal{L}_{\text{smooth}}^{\Omega} = &\int_{\Omega} \|\nabla g(\mathbf{x}) \|^2 d\mathbf{x}, \\ \mathcal{L}_{\text{align}} = & \int _{\mathcal{S}} \big( 1 - \langle \nabla g(\mathbf{x}), \textbf{v}_{\text{T}}(\mathbf{x}) \rangle ) \big) d \mathbf{x}.  \label{eq:align}
\end{align}
While the smooth Dirichlet loss aims to make the guidance field singularity-free in the interior (\rev{}{w.r.t., smooth energy $\mathcal{O}_{\text{smooth}}^{\Omega}$ in Eq.~\ref{eq:smoothInterior}), the align loss $\mathcal{L}_{\text{align}}$ ensures that the gradient of the final trained guidance field aligns with the previously optimal direction at the boundary (i.e., minimizing $\mathcal{O}_{\text{align}}$ in Eq.~\ref{eq:alignObjective})}.

When training for guidance INF, the minimum principal curvature direction $\mathbf{d}_{\kappa}(\mathbf{x})$ is used as the initial guess for $\nabla g(\mathbf{x})$. Using curvature direction, analytically computed from the SDF~\cite{goldman2005curvature}, offers several advantages. It naturally aligns with the local tangent plane (i.e., $\mathbf{d}_{\kappa}(\mathbf{x}) \subseteq \mathbf{T}_{\mathbf{x}}$), satisfies the support-free constraint (since the support-free angle is always $90^\circ$), and ensures that $\mathcal{L}_{\text{align}} = 0$. In our implementation, the warm-up training loss is defined as follows:
\begin{equation}
    \mathcal{L}_{\text{warm-up}}^{\text{guide}} = \mathcal{L}_{\text{coll}}^{\Omega_b} + \omega_{\kappa}\int_{\mathcal{S}} (1-\langle \nabla g(\mathbf{x}) , \mathbf{d}_{\kappa}(\mathbf{x}) \rangle ) d\mathbf{x}, \label{eq:warmup-guide}
\end{equation}
where the Neumann boundary condition $\mathcal{L}_{\text{coll}}^{\Omega_b}$ is also included. Similar strategy is also reported in references for field optimization on both discrete settings~\cite {jakob2015instant, vaxman2016directional} and neural-based solutions~\cite {dielen2021learning,dong2024neurcross}. Ablation studies on the inclusion of the initial guess selection and two-step field training are presented in Sec.~\ref{subsec:guideAblation}, demonstrating the effectiveness of the loss design in balancing different fabrication-oriented objectives to find optimum guidance field.

\begin{figure*}[!t]
    \centering
    \includegraphics[width=1.0\linewidth]{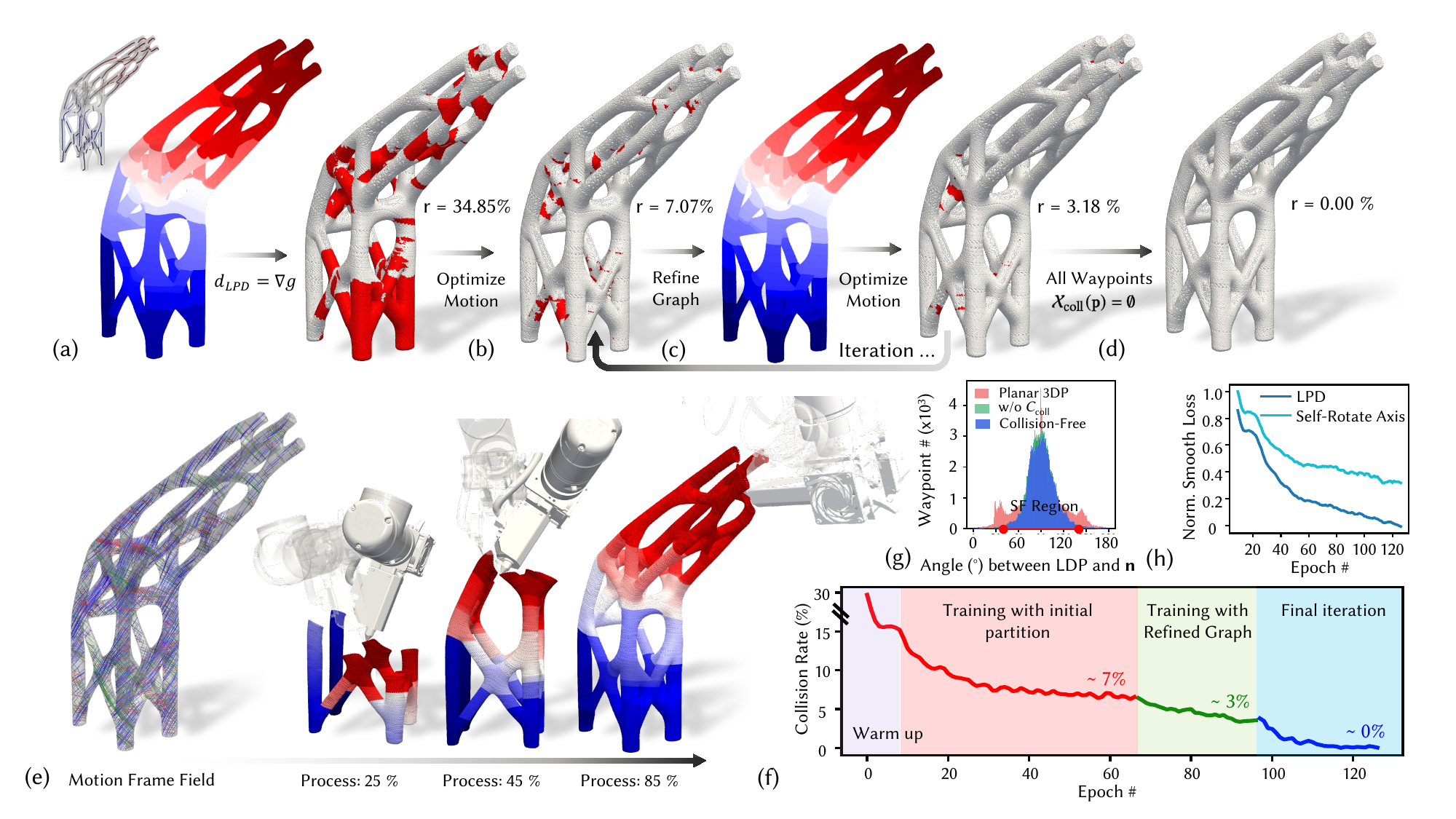}
    \caption{(a) For the Satellite Bracket model with complex topology, initial partitioning with a Reeb graph maximizes print continuity, but directly using $\nabla g$ as the LPD results in a high collision rate. (b) Partitioning constraints limit the solution space, and motion optimization alone reduces but does not eliminate collisions.
(c) Iterative refinement of partitions and motion optimization further lowers the collision rate.
(d) The process converges to a collision-free state after multiple iterations.
(e) Visualization of the final optimized motion frame field (streamlines) and key printing frames.
(f) The collision rate over training epochs shows that, after several iterations, the process converges to a collision-free state. (g) The adjusted LPD enables support-free printing with optimized $\mathcal{L}_{\text{SF}}^{\text{motion}}$ (Eq.\ref{eq:SF_motion}). (h) Training convergence curves for the smoothness of both the LPD and self-rotation axis (Eq.\ref{eq:smoothMotionLoss}).
}
    \label{fig_collision}
\end{figure*}

\subsection{Infill Field}  

The infill guidance fields $\psi$ control the geometry of the infill structure for solid printing. In particular, the growing direction and density of infill is controlled by $\nabla \psi (\mathbf{x})$ and its norm $\| \nabla \psi (\mathbf{x}) \|$, respectively. The loss used to optimize infill INF is set as
\begin{equation}
 \mathcal{L}_\mathrm{infill} = \omega_{\text{density}} \mathcal{L}_{\text{density}} + 
 \omega_{\text{SF}}^{\text{infill}} \mathcal{L}_{\text{SF}}^{\text{infill}} + \omega_{\text{smooth}}^{\text{infill}} \mathcal{L}_{\text{smooth}}^{\text{infill}}.
\end{equation}
where
\begin{align}
    \mathcal{L}_{\text{density}} &= \int_{\Omega} \big (\| \nabla \psi(\textbf{x}) \| - \rho(\mathbf{x}) \big) ^2 d\textbf{x}, \label{eq:densityLoss} \\
\mathcal{L}_{\text{smooth}}^{\text{infill}} &= \int_{\Omega} \| \nabla \psi(\textbf{x}) \|^2 d\textbf{x}, \\
    \mathcal{L}_{\text{SF}}^{\text{infill}} &=   \int_{\Omega} \sigma \big( k_{\text{SF}} ( \big | \langle 
    \frac{\nabla g(\mathbf{x})}{\|\nabla g(\mathbf{x}) \|} , 
    \frac{\nabla \psi(\mathbf{x})}{\|\nabla \psi(\mathbf{x}) \|}  
    \rangle \big | - \sin\alpha )  \big) d \mathbf{x}. \label{eq:SF_infill} 
\end{align}
%\Js{Interior $\Phi$ in equations is altered to $\Omega$.}
Where the density function $\rho(\mathbf{x})$ is also a spatial scalar field that controls the local density of the infill. If the density is constant across space (i.e., $\rho(\mathbf{x}) = c$), regular infill will be generated. Examples using variable density can be found in Fig.~\ref{teaser}(d), Fig.~\ref{fig_pipeline}(e), and Fig.~\ref{fig_infill}(c) - where local feature size detected by SDF is used as $\rho$. Note that the support-free loss of infill structure is determined w.r.t., $\nabla g$, and the singularity-free solution is also encouraged by minimizing smooth loss $\mathcal{L}_{\text{smooth}}^{\text{infill}}$. 
In the training process, a reference direction $\mathbf{d}_{\text{ref}}$ controlled by angle $\beta$ can be selected to control the direction of infill. The initial direction $\mathbf{d}_{\text{int}}(\beta,~\mathbf{d}_{\text{ref}},~\textbf{x})$ for $\nabla \psi$ is expressed as %\Js{Warm up means first used in training but then not use, but direction loss is always applied for different patterns, not same as curvature initialization.}
\vspace{5px}
\begin{equation*}
     \cos \beta~\frac{\nabla g(\textbf{x})}{\|\nabla g(\textbf{x})\|}\times \mathbf{d}_{\text{ref}} +  
    \sin \beta~\frac{\nabla g(\textbf{x})}{\|\nabla g(\textbf{x})\|}\times(\frac{\nabla g(\textbf{x})}{\|\nabla g(\textbf{x})\|}\times \mathbf{d}_{\text{ref}}),
\end{equation*}
\vspace{5px} which is always perpendicular to $\nabla g$ to ensure support-free constraint, and we set the reference loss for infill training as 
\begin{equation}
    \mathcal{L}_{\text{ref}} =  \int_{\Omega} (1-\langle \frac{\nabla \psi(\textbf{x})}{\|\nabla \psi(\textbf{x})\|} , \mathbf{d}_{\text{int}}(\beta,~\mathbf{d}_{\text{ref}},~\textbf{x}) \rangle ) d\mathbf{x},
\end{equation}
By varying angle $\beta$ and $\textbf{d}_{\text{ref}}$, diverse infill patterns can be generated by training multiple guidance fields $\{\psi_1,\psi_2,...\psi_k\}$. Results for different infill patterns, including cross, diamond, and hexagonal, are shown in Fig.~\ref{fig_infill}. Details on generating the waypoint $\mathcal{P}$ by discretizing the implicitly defined toolpath are presented in Sec.~\ref {sec_toolpath}.

\subsection{Iterative-based Optimization for Printing Sequence and Motion Frame Fields}
\label{subsec:motionTraining}

%\Guoxin{working here now --- May 7th}
%\Guoxin{why we need a continuous search for sequence field: for differentiable collision response as training. - this need to be highlight before.}
%Thanks to the alignment between material growing on surface and inertial by optimizing $\mathcal{L}_{\text{align}}$ defines in Eq.~\ref{eq:align}, 

As discussed in Sec.~\ref{subsec:motionObjective}, to ensure printing continuity, the model is partitioned into connected regions and encoded by a piecewise-constant function $\ell$. For each waypoint $\mathbf{p}$, the printing sequence can be defined as $\mathcal{T}(\textbf{x}) = g(\textbf{x}) + \ell(\textbf{x})$, with  $\ell(\textbf{x})$ trained with Cross-entropy loss from the partition data generated with Reeb-Graph. With the printing sequence field $\mathcal{T}$ being defined, the loss used to optimize printing motion as a quaternion field $\mathbf{q}$ that ensures a collision-free printing process with smooth machine motion is defined as

\begin{equation}
    \mathcal{L}_{\text{motion}} = \omega_{\text{coll}} \mathcal{L}_{\text{coll}} + \omega_{\text{SF}}^\text{motion} \mathcal{L}_{\text{SF}}^\text{motion} + \omega_{\text{smooth}}^{\text{motion}}\mathcal{L}_{\text{smooth}}^{\text{motion}}
\end{equation}
where
\begin{align}    
    \mathcal{L}_{\text{coll}} = &\sum_{\mathbf{p}\in \mathcal{P}} \sum_{\mathbf{x}\in \mathcal{X}_{\text{coll}}(\mathbf{p})} -\phi_{\mathcal{T}}(\mathbf{p}, \mathbf{x}), \\
    \mathcal{L}_{\text{SF}}^\text{motion} = &  \sum_{\mathbf{p} \in\mathcal{S}} \sigma \big( k_{\text{SF}} (\big | \langle \mathbf{R}(\mathbf{q}(\mathbf{p}))\cdot \textbf{z}, \nabla \phi (\mathbf{x})  \rangle \big |  - \sin\alpha ) \big) , \label{eq:SF_motion}\\ \mathcal{L}_{\text{smooth}}^{\text{motion}} = &\sum_{\mathbf{p}\in \mathcal{P}} 
    \big(
    \underbrace{ \| \nabla  (\mathbf{R}(\mathbf{q}(\mathbf{p}))\cdot \textbf{z}) \|^2 }_{\text{LPD}} + 
    \underbrace{ \| \nabla (\mathbf{R}(\mathbf{q}(\mathbf{p}))\cdot \textbf{x}) \|^2}_{\text{self-rotation axis}}  
    \big ). \label{eq:smoothMotionLoss}
\end{align}
%\Js{this should be $\nabla$ instead of $\nabla \cdot$} \\
The set $\mathcal{X}_{\text{coll}}(\mathbf{p}) = \{\mathbf{x} \in \mathcal{M}_{e}(\mathbf{p})~|~\phi_{\mathcal{T}}(\mathbf{p},\mathbf{x}) < 0\}$ includes all collision points in the printing setup. The collision loss $\mathcal{L}_{\text{coll}}$ is computed as the sum of the distances from given points to the surface of the partially printed object at each sequence step $\mathcal{T}(\mathbf{p})$. Specifically, it is evaluated as $-\phi_{\mathcal{T}}(\mathbf{p}, \mathbf{x})$, as detailed in Sec.~\ref{subsec:Time-SDF}.

The support-free loss is defined using LPD, which is obtained by transforming $\mathbf{z}$ with the rotation matrix $\mathbf{R}$ derived from the quaternion $\mathbf{q}(\mathbf{p})$. Notably, the motion smoothness loss $\mathcal{L}_{\text{smooth}}$ is evaluated separately on the LPD and its orthogonal axis, which controls nozzle self-rotation. Compared with directly enforcing smoothness on the quaternion, our approach improves convergence by finding a smooth motion frame, as neural-based quaternions can result in multiple solutions for the same motion (see ablation study and discussion in Sec.~\ref{subsec:guideAblation}).

As discussed in Sec.~\ref{subsec:shortDiscuss}, the initial printing sequence $\mathcal{T}$ is determined with the help of Reeb graph with maximized continuity in the printing process. \rev{}{Here, the graph is constructed following a similar method as presented in~\cite{zhong2023continuous}, where the center of each connected iso-contour region (extracted with optimized scalar field $g$) generates a node of the graph. Neighboring regions, whose center-to-center distance is below a threshold (set as $2\times$ average layer height), are connected as edges.} However, this approach can significantly restrict the collision-free solution space, so subsequent optimization of the motion field may still fail to satisfy the collision-free constraint. For example, as shown in Fig.~\ref{fig_collision} for the satellite bracket model, minimizing $\mathcal{L}_{\text{coll}}$ with the initial $\mathcal{T}$ significantly reduces the collision rate from $34.85\%$ to $7.07\%$, but still cannot converge into zero due to the lack of a feasible solution that also ensures support-free and smooth motion. We address this issue by iteratively refining the partition in critical regions and retraining both printing sequence $\mathcal{T}$ and motion quaternion field $\mathbf{q}$, until a strictly collision-free printing motion is achieved for all waypoints. An illustration of this process is shown in Fig.~\ref{fig_collision}(b), where the final result satisfies both $\mathcal{C}_{\text{coll}}$ and $\mathcal{C}_{\text{SF}}$, with motion smoothness optimized along both axes.

Although this iterative process will sacrifice optimal printing continuity, we claim that manufacturability remains the top priority for successful printing. Notably, the convergence of this iterative optimization is guaranteed — in the worst case, the process converges to $\mathcal{T}(\mathbf{x}) = g(\mathbf{x})$, where large discontinuities are introduced to ensure a collision-free solution. Based on our test, the relative size of the model to the printing setup greatly influences the number of iterations required (see Sec.~\ref{subsec:motionPlanResult} for a detailed discussion). For all challenging models tested in this work, the proposed pipeline effectively plans smooth collision-free motions while also maintaining support-free printing.

%% file: tex/implementation.tex
\section{Implementation Details}

In this section, we first describe how to construct the time-varying SDF of a growing object via field intersection, enabling differentiable collision loss evaluation for motion planning. Next, we outline the parallel computing process for discretizing implicitly defined toolpaths using field projection, efficiently generating high-precision printing waypoints. Finally, we present the network architecture and learning process for INFs.

\begin{figure}[!t]
    \centering
    \includegraphics[width=1.0\linewidth]{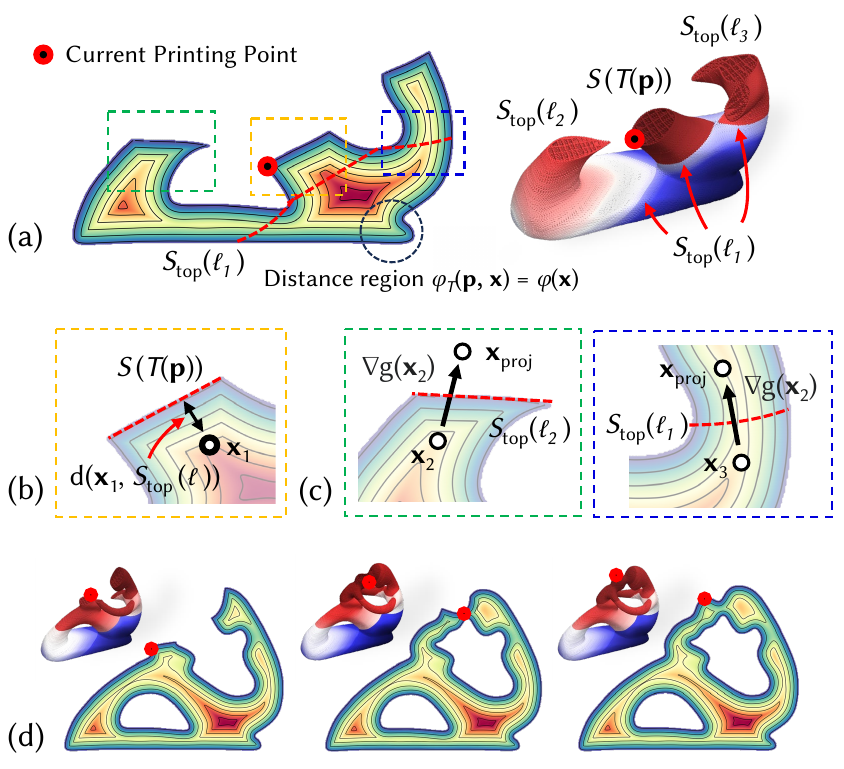}
    \caption{Illustration of the time-varying SDF during material deposition:
(a) As printing progresses to waypoint $\mathbf{p}$, the current working surface and the already printed region are identified using the printing sequence $\mathcal{T}(\mathbf{p})$ and $g(\ell)$, respectively.
(b) Zoomed-in view around $\mathbf{p}$, showing the point-to-surface distance to $\mathcal{S}(\mathcal{T}(\mathbf{p}))$.
(c) Points where $d(\mathbf{x}, \mathcal{S}_{\text{top}}(\ell)) < |\phi(\mathbf{x})|$ are highlighted. If the projected point $\mathbf{p}_{\text{proj}}$ falls inside the already printed region (left), $\phi_{\mathcal{T}}$ is updated; otherwise $\phi_{\mathcal{T}}$ is not updated (right).
(d) Key frames of the constructed time-varying SDF. While the distance is approximated using Eq.~\ref{eq:approxDistance} to effectively compute collision loss $\mathcal{L}_{\text{coll}}$, we consistently ensure correct sign evaluation for $\phi_\mathcal{T}$, supporting strict collision-free planning.
}
\label{fig_SDFWarp}
\end{figure}

\subsection{Construction of Time-Varying SDF and Collision Loss} \label{subsec:Time-SDF}

To evaluate global collision through the printing process, the shape of the dynamically growing object when the printing comes to waypoint $\mathbf{p}$ needs to be constructed as a time-varying SDF for the already printed region (see Fig.~\ref{fig_SDFWarp}).
%At a given waypoint $\mathbf{p}$ in the printing sequence $\mathcal{T}(\mathbf{p})$, we must accurately represent the geometry of the partially printed object. 
%Instead of retraining time-varying SDF at each printing step \Aoran{just a note: this might be challenged by the reviewer. As g(x) and l(x) are fixed, you may train a $\phi(\bm{x}, \mathcal{T})$ using a slightly larger network.}
%and make the loss function $\mathcal{L}_{\text{coll}}$ not differentiable. \Aoran{I think $\mathcal{L}_{\text{coll}}$ is still differentiable in this case.} 
As discussed in Sec.~\ref{subsec:shortDiscuss}, rather than training an end-to-end time-dependent SDF network, which can lead to inaccuracies in collision checking. In this work, we propose a training-free method using direct field interpolation to construct the time-varying SDF that guarantees exact collision checking.

%and can efficiently approximate the collision loss $\mathcal{L}_{\text{coll}}$ for motion planning.

We first consider those printed points that locates in the same partition level of the current printing waypoint, i.e., $\{\mathbf{x} \mid \ell(\mathbf{p}) = \ell(\mathbf{x}) \land \phi(\mathbf{x}) \le 0 \land g(\mathbf{x})\le g(\mathbf{p})\}$ - see Fig.~\ref{fig_SDFWarp}(b). The time-varying SDF at these points can be computed by comparing the global SDF $\phi(\mathbf{x})$ with the distance from $\mathbf{x}$ to the surface $\mathcal{S}(\mathcal{T}(\mathbf{p})) := \{ \mathbf{x}\mid \mathcal{T}(\mathbf{x}) = \mathcal{T}(\mathbf{p}) \}$, which can be defined as
%the length of shortest path from $\mathbf{x}$ to the surface along the gradient direction 
\begin{equation} \label{eq:point-to-surface-dis}
    d(\mathbf{x}, \mathcal{S}(\mathcal{T}(\mathbf{p}))) = \min_{~\bf{c}\in\mathcal{S}(\mathcal{T}(\mathbf{p}))}\|\bf{x}-\bf{c}\|.
\end{equation}
If $d(\mathbf{x}, \mathcal{S}(\mathcal{T}(\mathbf{p}))) < -\phi(\mathbf{x})$, the time-varying SDF $\phi_{\mathcal{T}}(\mathbf{p},\mathbf{x})$ will be updated as $-d(\mathbf{x}, \mathcal{S}(\mathcal{T}(\mathbf{p})))$.
If the object only has one partition level, this simple formula can precisely derive the time-varying SDF for the whole internal region. 
% (in this case, the points $\textbf{x}$ always inside the model with $\phi(\textbf{x})<0$).
%\Aoran{Actually the shortest path may not along the gradient direction, the real distance is $\min_{\bf{p}\in\mathcal{S}(\mathcal{T}(\mathbf{p}))}\|\bf{x}-\bf{p}\|$. So I suggest still derive the approximation from the normal of the isosurface.}
% \vspace{-2px}
% \begin{equation*}
%     d(\mathbf{x}, \mathcal{S}(\mathcal{T}(\mathbf{p}))) = \int_{g(\mathbf{x})}^{g(\mathbf{p})} \frac{1}{\|\nabla g(s)\|} ds
% \end{equation*}
%\Guoxin{$\mathcal{B}$ can be expressed using matrix form. (but not adjacency matrix) - this matrix can be viulized in the figure.}

However, with graph-based partitioning used to ensure continuity in the printing process, multiple top surfaces with different printing sequence values can exist at the same time
% unfinished regions with different partition indices and individual top surfaces
- see right side of Fig.~\ref{fig_SDFWarp}(a). 
% The $d_{\mathbf{S}}$ computed in Eq.~\ref{eq:point-to-surface-dis} is only guaranteed to be accurate for the point near the current waypoint. 
In this case, we must find the distance from a query point $\mathbf{x}$ to its closet top surface.
% To more accurately compute the time-varying SDF for all points $\mathbf{x}$ in space, we must consider not only the distance to $\mathcal{S}(\mathcal{T}(\mathbf{p}))$, but also the distance to the top surfaces of all the already printed regions with higher $\ell$ than $\ell(\mathbf{x})$. 
Here we define a set $\mathcal{B}(\mathbf{p},\mathbf{x}) = \{ \ell~|~ \ell(\mathbf{x}) \leq \ell \leq \ell(\mathbf{p})\}$, representing the indices of all currently printed partitions that printed after $\mathbf{x}$. The closet top surface of $\mathbf{x}$ will belong to one of the partitions in $\mathcal{B}(\mathbf{p},\mathbf{x})$. For a partition in $\mathcal{B}(\mathbf{p},\mathbf{x})$ with index $\ell_k$, its highest value of the guidance field can be calculated as $g_k = \max \{ g(\mathbf{x}) \mid \ell(\mathbf{x}) = \ell_k \land \mathcal{T}(\mathbf{x}) \le \mathcal{T}(\mathbf{g})\}$, which defines its top surface (can be covered by other partitions) as $\mathcal{S}_{\text{top}}(\ell_k) := \{ \mathbf{x} \mid \mathcal{T}(\mathbf{x}) = g_k + \ell_k \}$.
% , and  the overall time-varying SDF is defined as:
% \begin{equation}
%     \phi_{\mathcal{T}}(\mathbf{p},\mathbf{x}) = \max \big\{  
%     \phi(\mathbf{x}),\,
%     \max_{\ell \in \mathcal{B}(\mathbf{p},\mathbf{x})} - d(\mathbf{x}, \mathcal{S}_{\text{top}}(\ell)) \big\}, 
%     \label{eq:PSDF}
% \end{equation}
%where the set $\mathcal{B}(\mathbf{p},\mathbf{x}) = \{ \ell~|~ \ell(\mathbf{x}) \leq \ell \leq \ell(\mathbf{p})\}$ includes all region indices that is smaller than $\mathbf{p}$ and larger than $\mathbf{\mathbf{x}}$. For the $k$-th region with index as $\ell_k$, the highest guidance field value is computed as $g_k = \max \{ g(\mathbf{x}) \mid \ell(\mathbf{x}) = \ell_k \}$, which defines the top surface as $\mathcal{S}_{\text{top}}(\ell_k) := \{ \mathbf{x} \mid g(\mathbf{x}) = g_k \}$. 
% Note that $\ell(\mathbf{p})$ always present the highest sequence $\mathcal{T}$ when printing at waypoint $\mathbf{p}$, and therefore $\mathcal{S}(\mathcal{T}(\mathbf{p})) = \mathcal{S}_{\text{top}}(\ell(\mathbf{p}))$ ?holds.
Note that $\mathcal{S}(\mathcal{T}(\mathbf{p})) = \mathcal{S}_{\text{top}}(\ell(\mathbf{p}))$ holds under this definition.

\begin{algorithm}[t]
\caption{Time-Varying SDF for evaluating $\mathcal{L}_{\text{coll}}$}
\LinesNumbered
\label{algorithm:SDF}
\KwIn{
    Waypoint set $\mathcal{P}$, SDF network $\mathcal{N}_{\text{SDF}}$, partition network $\mathcal{N}_{\text{part}}$, and guidance network $\mathcal{N}_{\text{guide}}$.
}
\KwOut{Time-varying SDF $\phi_{\mathcal{T}}$ and collision loss $\mathcal{L}_{\text{coll}}$.}

\textbf{set} $\mathcal{L}_{\text{coll}} = 0$;\

\ForPar{$\mathbf{p} \in \mathcal{P}$}{
    \tcp{Current printing step}
     $\ell(\mathbf{p})\gets \mathcal{N}_{\text{part}}(\mathbf{p})$, $\mathcal{T}(\mathbf{p}) \gets \mathcal{N}_{\text{guide}}(\mathbf{p}) + \ell (\mathbf{p})$

    \ForPar{$\mathbf{x} \in \mathcal{M}_e$}{
    \tcc{Transform $\mathcal{M}_e$ to current waypoint}
    $\mathbf{x} \gets \mathbf{R}(\mathbf{q(\mathbf{p})})\mathbf{x} + \mathbf{p}$  ;\

    $\phi(\mathbf{x}) \gets \mathcal{N}_{\text{SDF}}(\mathbf{x})$;

    % \If{$\phi(\mathbf{x}) > 0$}{$\phi_{\mathcal{T}}(\mathbf{p},\mathbf{x}) = 0$  \tcp*{collision-free point} 
    \If{$\phi(\mathbf{x}) > 0 \lor \mathcal{T}(\mathbf{x})>\mathcal{T}(\mathbf{p})$}{
    \textbf{continue}  \tcp*{collision-free point} 
    }

    $\phi_{\mathcal{T}}(\mathbf{p},\mathbf{x}) \gets \phi(\mathbf{x})$, $\ell(\mathbf{x})\gets \mathcal{N}_{\text{part}}(\mathbf{x})$ 
    
    $\mathcal{B}(\mathbf{p},\mathbf{x}) = \{ \ell~|~ \ell(\mathbf{x}) \leq \ell \leq \ell(\mathbf{p})\}$ \tcp*{Region index} 
    
    \ForEach{$\ell_k \in \mathcal{B}(\mathbf{p})$}{
        $g_k \gets \max\{g(\mathbf{x}) \mid \ell(\mathbf{x}) = \ell_k\}$ \;
       % $\mathcal{S}_{\text{top}}(\ell_k) \gets \{\mathbf{x} \mid g(\mathbf{x}) = g_k\}$ \;
        Approximiate $d_k = d(\mathbf{p}, \mathcal{S}_{\text{top}}(\ell_k))$ by Eq.~\ref{eq:approxDistance} \;
       \If{$d_k<|\phi_{\mathcal{T}}(\mathbf{p},\mathbf{x})|$}{
        $\mathbf{x}_{\text{proj}} \gets \mathbf{x} + 2 d_k \nabla g(\mathbf{x})/ \|\nabla g(\mathbf{x})\|$ \;
        \If{$\ell(\mathbf{x}_{\text{proj}}) \notin \mathcal{B}(\mathbf{p}, \mathbf{x})$} {
            $\phi_{\mathcal{T}}(\mathbf{p},\mathbf{x}) \gets -d_k$ \tcp*{Update $\phi_{\mathcal{T}}$}
        }
       }
    }

$\mathcal{L}_{\text{coll}} \gets \mathcal{L}_{\text{coll}} + |\phi_{\mathcal{T}}(\mathbf{p,\mathbf{x}})|$
    
    }
}
\Return{$\mathcal{L}_{\text{coll}}$}
\end{algorithm}

As illustrated in Fig.~\ref{fig_SDFWarp}(c), when a point is detected $d(\mathbf{x}, \mathcal{S}_{\text{top}}(\ell_k)) < -\phi(\mathbf{x})$ with a top surface $\mathcal{S}_{\text{top}}(\ell_k)$, we perform an additional check to ensure proper merging of time-varying SDF between regions with different partition indices. Following a similar SDF operation procedure proposed in~\cite{marschner2023constructive}, we use gradient tracing to find a corresponding projected query point $\mathbf{x}_{\text{proj}} = \mathbf{x} + 2 d_k \nabla g(\mathbf{x}) / \|\nabla g(\mathbf{x})\|$. Whether this projected point lies within a region that has already been printed (i.e., if $\ell(\mathbf{x}_{\text{proj}}) \in \mathcal{B}(\mathbf{p},\mathbf{x})$) determines whether the SDF value at $\mathbf{x}$ should be replaced by $-d(\mathbf{x}, \mathcal{S}_{\text{top}}(\ell_k))$.
%If the projected point is inside the printed region, we do not update the SDF value at $\mathbf{x}$ for $\phi_{\mathcal{T}}$. 
This process ensures that transitions between printed regions are handled smoothly and gaps are avoided (see Fig.~\ref{fig_SDFWarp}(d) for a result).
%\Guoxin{more discussion on this limitation is needed.} \Aoran{This will be very hard to understand if you don't point out that this is for finding the correct branch above the query point. But if you point it out, counter examples will be suddenly available} 
The pseudo-code of evaluation of time-varying SDF and $\mathcal{L}_{\text{coll}}$ can be found in Algorithm~\ref{algorithm:SDF}.

It is worth mentioning that, when evaluating the point-to-surface distance $d(\mathbf{x},\mathcal{S}_{\text{top}}(\ell_{k}))$, we do not solve the projection problems Eq.~\ref{eq:point-to-surface-dis} for every point in the space, due to the lack of an explicit surface definition. Instead, we use the following approximation:
\begin{equation}
    d(\mathbf{x}, \mathcal{S}_{\text{top}}(\ell_{k})) \approx (g_k-g(\mathbf{x}))/\|\nabla g(\mathbf{x})\|. \label{eq:approxDistance}
\end{equation}
to achieve an efficient and differentiable form for collision loss $\mathcal{L}_{\text{coll}}$. 
% This approximation will not bring the issue of collision checking as the sign of $\phi_{\mathcal{T}}$ is always kept correct. 
This approximation will not affect our exact collision checking, as we directly use $\phi(\mathbf{x})$ and $\mathcal{T}(\mathbf{x})$ to identify all collisions.
For point $\mathbf{x}$ far away from the current printing surface $\mathcal{S}(\mathcal{T}(\mathbf{p}))$, the error in the approximation increases, but the time-varying SDF generally converges to the global SDF: $\phi(\mathbf{x}) = \phi_{\mathcal{T}}(\mathbf{x})$ for distant points (see the bottom part of fertility model in Fig.~\ref{fig_SDFWarp}(a)). With the help of this approximation, the gradient of collision loss w.r.t., network parameter $\theta$ of the motion INF can be expressed by the chain rule as
\begin{equation}
    \frac{\partial \mathcal{L}_{\text{coll}}}{\partial \theta}(\theta)
= - \sum_{\mathbf{p} \in \mathcal{P}}
    \sum_{\mathbf{x} \in \mathcal{X}_{\text{coll}}(\mathbf{p})}
        \left[\frac{\partial \phi_{\mathcal{T}}}{\partial \mathbf{x}}(\mathbf{x},\mathbf{p})\right]^{\rm{T}}
        \left[\frac{\partial \mathbf{x}}{\partial \mathbf{q}}(\mathbf{p};\theta)\right]
        \left[\frac{\partial \mathbf{q}}{\partial \theta}(\mathbf{p};\theta)\right]
\end{equation}
% where for the first term if remark Eq.~\ref{eq:PSDF} as $\max \{f_1,~f_2\}$, we have
where for the first term we have the two following cases, depending on the updating process in Algorithm~\ref{algorithm:SDF}:
% \begin{equation}
% \begin{split}
%  %\frac{\partial \phi_{\text{P-SDF}}}{\partial \mathbf{x}}
% \frac{\partial f_1}{\partial \mathbf{x}} & = \nabla \phi(\mathbf{x}) 
% \\
% \frac{\partial f_2}{\partial \mathbf{x}} & = -\frac{\nabla g(\mathbf{x})}{\left\| \nabla g(\mathbf{x}) \right\|}
% +
% \frac{g(\mathbf{x}) }{\left\| \nabla g(\mathbf{x}) \right\|^3}
% \,
% \nabla g(\mathbf{x})^{T} \nabla^2 g(\mathbf{x}) \nabla g(\mathbf{x}).
% \end{split}
% \end{equation}
\begin{equation}
\frac{\partial \phi_{\mathcal{T}}}{\partial \mathbf{x}}=
\begin{cases}
\nabla \phi(\mathbf{x}), & \phi_{\mathcal{T}}=\phi,\\
-\frac{\nabla g(\mathbf{x})}{\left\| \nabla g(\mathbf{x}) \right\|}
+
\frac{g(\mathbf{x}) }{\left\| \nabla g(\mathbf{x}) \right\|^3}
\,
\nabla g(\mathbf{x})^{T} \nabla^2 g(\mathbf{x}) \nabla g(\mathbf{x}),  & \text{others}.
\end{cases}
\end{equation}
%Moreover, since $\nabla d = \nabla \mathcal{T} = \nabla g$, the pipeline remains differentiable, as discussed in the following section. 
In our partial training scheme, the second-order term in the gradient expression is omitted, resulting in
$\partial \phi_{\mathcal{T}} / \partial \mathbf{x} \approx - \nabla g(\mathbf{x})/\| \nabla g(\mathbf{x}) \|$ for the second case. This simplification makes the gradient
$\partial \phi_{\mathcal{T}}/\partial \mathbf{x}$
align with the fastest direction of guiding the collision point out of the printed model. The exact direction depends on which distance metric is currently dominant in the time-varying SDF evaluation. Such gradient-based approaches are widely adopted in robotics for collision response and avoidance~\cite{mirrazavi2018unified}, providing a principled and computationally efficient means to generate corrective motions and maintain safe configurations during printing.

%\subsection{Differentiable Pipeline for Collision Response}

%All loss formulation presented in Sec.~\ref{} for field training are differenitiable, and gradient-based solver~\cite{} is used. Here we give the major analysis for the collision loss to proof the efficiency of motion planning process. 

\subsection{Parallel Computing for Waypoint Generation}
\label{sec_toolpath}

\begin{figure}[t]
    \centering
    \includegraphics[width=1.0\linewidth]{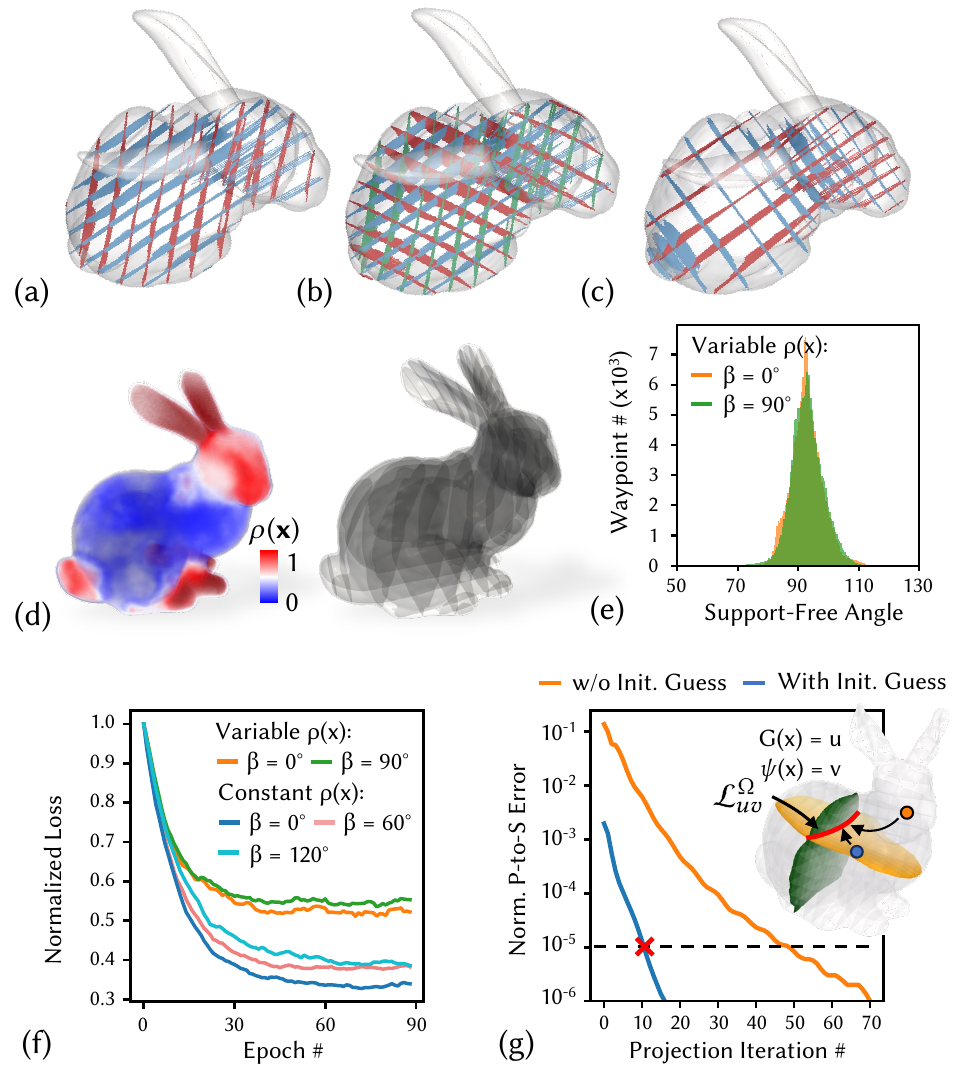}
    \caption{Different types of infill patterns can be achieved by efficiently training ${\psi}$ and computing the corresponding toolpaths. (a) Diamond infill pattern, created using two infill fields with an angular difference of $\beta = 45^\circ$. (b) Hexagonal infill pattern, using three infill fields with $\beta = 60^\circ$. (c) Variable density infill generated by incorporating (d) geometry-feature-defined density $\rho$ into $\mathcal{L}_{\text{density}}$ (Eq.~\ref{eq:densityLoss}) during training. (e) All infill structures are guaranteed to be support-free through optimization of the support-free loss $\mathcal{L}_{\text{SF}}^{\text{infill}}$ (Eq.~\ref{eq:SF_infill}). (f) All infill fields converge rapidly in optimization. (g) Comparison of convergence with (blue curve) and without (orange curve) grid-based initialization as the initial guess for waypoint discretization by solving field projection problem (Eq.~\ref{eq:waypointProject}).
}
    \label{fig_infill}
\end{figure}

%\Guoxin{Naive solution (don't work) from one point at $\mathcal{L}$ search with direction $\mathbf{t}(\mathbf{x}) = \nabla g(\mathbf{x}) \times \nabla \psi(\mathbf{x}),$ to a lenght into next point. unlike SDF that this search is correct, issue with two SDF, where search from direction will not guanatee the point satisfies both case, and also the connectivity of these waypoint is not ensured.}

%\Guoxin{final algorithm: marching cube (connectivity is guanatee) - for each point ADMM (to precise project the point into $\mathcal{L}$) - arc-length parameterization - uniform sampling based on given distance (physically defined).}

\rev{for shells and  for infill (see Fig.~\ref{fig_infill}(g))}{With the help of INFs representation, the toolpath (w.r.t., iso-values $u, v$) for shell and solid structures is implicitly defined as iso-curves computed by field intersection:
\begin{equation*}
\begin{aligned}
    &\text{Shell:} ~~\mathcal{L}_{u}^{\mathcal{S}} ~= \{\mathbf{x} \mid g(\mathbf{x}) = u, \phi(\mathbf{x}) = 0\} \\
    &\text{Infill:} ~~\mathcal{L}_{uv}^{\Omega} = \{\mathbf{x} \mid g(\mathbf{x}) = u, \psi(\mathbf{x}) = v, \phi(\mathbf{x}) < 0\}
\end{aligned}
\end{equation*} }
As shown in Fig.~\ref{fig_infill}(g), the practical fabrication requires discretizing these continuous curves into a sequence of waypoints with consistent spacing $\Delta l$ and correct order for printing (here, without loss of generality we illustrate the method for infill toolpath $\mathcal{L}_{uv}^{\Omega}$). An intuitive approach is to trace along the toolpath by starting from an initial point $\mathbf{x}_0 \in \mathcal{L}$ and iteratively advancing with the tangent direction $\mathbf{t}(\mathbf{x}) = \nabla g(\mathbf{x}) \times \nabla \psi(\mathbf{x})$, seeking next waypoints as $\mathbf{x}_{\text{next}} = \Delta l \cdot\mathbf{t}(\mathbf{x}) + \mathbf{x}_0$. 
%\Aoran{Oh you mean this tracing, this tracing will not work even for a single SDF} \Aoran{This tracing equals to an explict Euler integration of ODE $\dot{\bf{x}}=\bf{t}(\bf{x})$ with timestep $\Delta l$. It's inaccuracy is well observed in many areas such as fluid simulation.} While this method is effective for a single SDF~\cite{} \Guoxin{TBA}, for toolpath as intersections of two implicit fields it does not guarantee that subsequent points satisfy both constraints - 
This tracing is equivalent to an explicit Euler integration of ODE $\dot{\bf{x}}=\bf{t}(\bf{x})$ with timestep $\Delta l$. 
%Its inaccuracy is well observed in many areas such as fluid simulation. 
Numerical errors can cause drift from the intersection manifold, and the process is difficult to compute in a parallel manner.

To address these challenges, we propose a robust, parallelizable method for waypoint generation that combines the strengths of grid-based and optimization-based techniques. First, we apply a Marching Cubes-style algorithm~\cite{lorensen1998marching} to extract the isosurface $g(\mathbf{x}) = u$ and then identify candidate intersection points with $\psi(\mathbf{x}) = v$ to find the grid-based solution of the isocurve. This grid-based approach ensures that the initial set of candidates is well-connected. 
%and covers the entire intersection topology \Aoran{I think the two descriptions are overlapped}. 
However, due to discretization, these candidates do not precisely satisfy both implicit constraints. Therefore, each candidate point $\mathbf{x}_0$ is refined by solving the following projection problem:
\begin{equation} \label{eq:waypointProject}
    \arg\min_{\mathbf{x}} \frac{1}{2} \|\mathbf{x} - \mathbf{x}_0\|^2 \quad \text{s.t.} ~g(\mathbf{x}) = u,\ \psi(\mathbf{x}) = v,
\end{equation}
This numerical projection step is highly parallelizable, as each waypoint can be processed independently. After projection, the set of waypoints is parameterized by arc length, and uniform resampling with the target spacing $\Delta l$ is performed to ensure consistent physical intervals between waypoints. The projection and resampling are done iteratively to make sure final waypoints are sufficiently close to the manifold and uniformly distributed. 

Based on our tests and as illustrated in Fig.~\ref{fig_infill}(g), grid-based initialization ensures rapid convergence of the optimization. \rev{}{Using the Augmented Lagrangian Method solver}, the process typically converges within ten iterations, achieving a normalized error of less than $10^{-5}$ (i.e., $0.01~mm$ for a 1-meter model). The convergence of the problem in Eq.~\ref{eq:waypointProject} is further guaranteed by the smoothness of both the guidance and infill fields, which provide reliable gradient directions for optimization. As also presented on the right of Fig.~\ref{teaser}(h) and Fig.~\ref{fig:bubble}(d), applying the waypoint projection step significantly improves the precision of surface waypoints with respect to the model surface, highlighting the advantage of our INF-based approach over existing explicit-representation-based solutions.

\subsection{Network Architecture and Sampling/Training Details}
\label{subsec:learningDetail}

We implement our field optimization with the PyTorch framework. All neural fields, except the sequence field, are represented using a neural network architecture based on SIREN~\cite{sitzmann2020implicit}. Specifically, each field network consists of 5 layers with hidden layer size 256, and employs periodic activation functions throughout. The Adam optimizer~\cite{kingma2014adam} is used for training, with a learning rate of $1 \times 10^{-4}$. The partition field is represented by a \textit{multi-layer perceptron} (MLP) comprising 3 layers with hidden layer size 256, and takes a 4-dimensional input $(\mathbf{x}, g(\mathbf{x})) \in \mathbb{R}^4$. The output dimension corresponds to the number of sequence levels determined by model partitioning. ReLU activations are used in the sequence field MLP, and sequence labels $\ell \in \mathbb{R}$ are obtained as the index of the maximum logit after applying Softmax. This MLP is trained using the Adam optimizer with a learning rate of $1 \times 10^{-3}$.

The model domain for all neural fields is a normalized cube, $\mathbf{x} \in [-1, 1]^3$, which is mapped to the physical dimensions of each fabricated object through linear scaling. The total number of training points across models of different sizes is directly proportional to their bounding box volumes. We find that this strategy performs stable convergence and accurate field representations, regardless of the object’s scale. \rev{}{During training, we apply different sampling strategies that adapt the training points across various INFs. Specifically, curvature-based non-uniform surface sampling \cite{novello2022exploring}, combined with uniform off-surface sampling, is used for SDF training. Across all tested models, the average SDF fitting error is less than $2.5 \times 10^{-4}$ with this sampling method.}

\rev{}{Additionally, for the guidance field training, both curvature-based and feature-size-based surface sampling methods are applied, while the interior regions (bounded by the SDF) are sampled based on the feature size. This ensures that regions with smaller volumes and higher complexity receive a higher sampling density. Notably, the first-stage training (i.e., minimizing $\mathcal{L_{\text{guide}}^{\mathcal{S}}}$ in Eq.~\ref{eq:guideSloss}) uses only surface points, while the second stage incorporates both. Similarly, the infill fields also use feature-size-based sampling. Finally, quaternion field training for collision-free motion uses 100k randomly sampled waypoints per epoch to enable fast training, while the exact collision ratio is computed for all waypoints based on the constructed TV-SDF (i.e., $\mathcal{C}_{\text{coll}} = 0$ for all waypoints being satisfied). Early stopping is triggered only when the collision ratio converges to zero, ensuring a restricted global collision-free motion sequence is found.}

% we employ a sampling strategy that adapts the number of training points to the size and complexity of the target model - i.e., larger models are assigned proportionally more sampling points to ensure adequate coverage of geometric details and field variations. This adaptive sampling helps maintain consistent learning quality across models of different sizes and prevents underfitting in larger domains. For each training epoch, points are randomly sampled within the normalized domain, with density proportional to the model’s bounding box volume. This ensures that the learned fields faithfully represent fine features and global structures. We observe that this approach enables stable convergence and high-fidelity field representations regardless of the object’s scale. The impact of model size and sampling density on training efficiency and accuracy is discussed in the next section.

\begin{figure}[!t]
    \centering
    \includegraphics[width=1.0\linewidth]{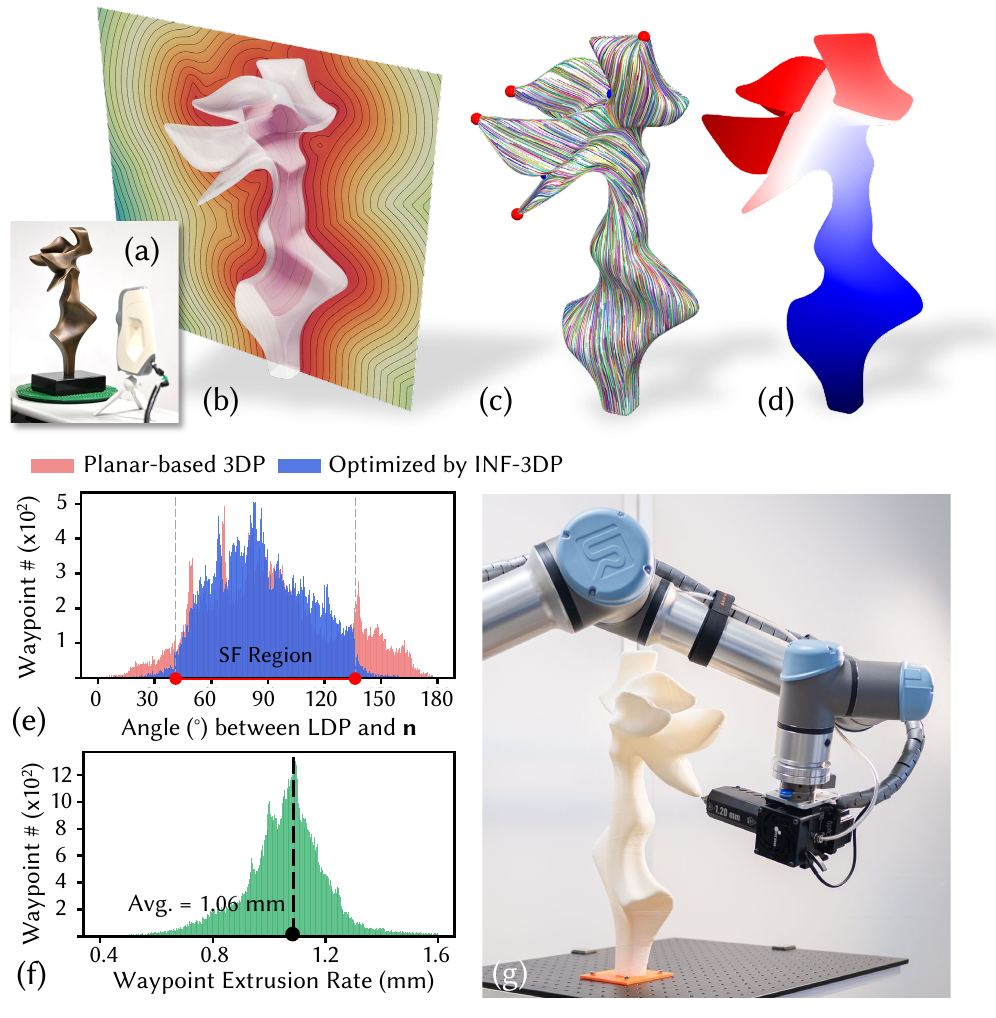}
    \caption{Computation results and fabrication process for the statue-goddess model: (a, b) The SDF is directly trained on a high-resolution point cloud captured by a 3D scanner. (c) The optimized guidance field ensures all singularities converge to locations that minimally affect surface quality. (d) The final toolpath and printing sequence are designed for as-continuous-as-possible motion. (e) Compared to planar-based printing, our approach achieves support-free fabrication. (f) The local extrusion rate is controlled within the 2.5 mm nozzle’s capability. (g) Even at the most challenging bottom-to-top printing angles, our method enables support-free, collision-free, and high-quality prints, as demonstrated in the physical setup.}
    \label{fig_statue}
\end{figure}

%Other hyperparameters, including the weights of loss terms for different field optimizations and further training details, are provided in Sec.~\ref{sec:ablation} (Ablation Studies) and are further discussed in the result analysis section. 

%% file: tex/result.tex
\section{Results and Discussion}
    \label{sec:result}

This section presents both computational and physical fabrication results. We have conducted extensive experiments with the INF-3DP pipeline on graphics models (Bunny, Fertility, Light Bulb) and industrial models (Statue-Goddess, Satellite Bracket, TO-Connector) to demonstrate the effectiveness and generality of our method. Comparisons with existing approaches (mainly based on mesh representation) are also provided. The proposed INF-3DP computational pipeline is implemented in Python for field training, waypoint generation, and motion planning. All computational experiments, including comparisons with existing methods, are conducted on a desktop PC equipped with an Intel Core i9-14900K CPU (24 cores @ 3.2 GHz), an NVIDIA RTX 4090 GPU, and 64 GB RAM, running Ubuntu 22.04.4 LTS. \rev{}{The implementation of this work is open-source and publicly available\footnote{https://github.com/Qjiasheng/INF-3DP}}.

\subsection{Computational Results and Comparison}

%\subsubsection{Examples and Results in Field Optimization}

\begin{figure*}[t]
    \centering
    \includegraphics[width=1.0\linewidth]{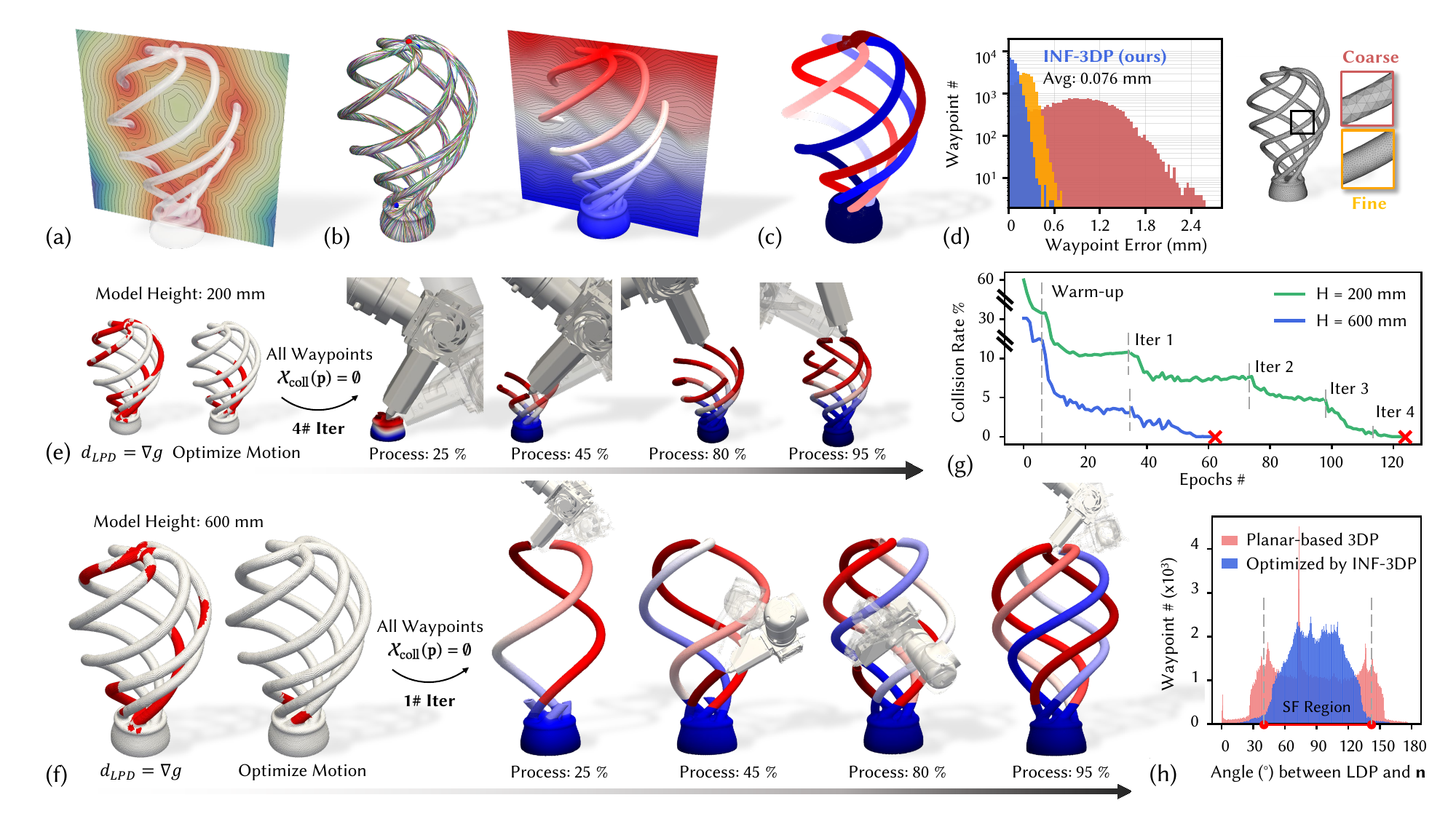}
    \caption{Computational results for toolpath generation and motion planning on the Light Bulb model.
(a) SDF trained using vertices from a dense mesh (average edge length $0.3\%$ of bounding box size, mesh shown in (d)).
(b) Optimized guidance field for both 2-manifold and 3-manifold domains.
(c) Initial toolpath and printing sequence from Reeb-graph partitioning, which leads to significant global collisions (highlighted in red)—even after motion optimization (see left of (e) and (f)).
(d) Comparison with mesh-based solutions, showing that the INF representation significantly reduces waypoint-to-surface error.
(e) and (f) Key steps of the printing process demonstrating final collision-free sequences for the 200 mm and 600 mm models, respectively.
(g) Convergence of collision rate through iterative printing sequence refinement and motion planning.
(h) Support-free printing is achieved with both the optimized toolpath and motion field.}
    \label{fig:bubble}
\end{figure*}

%\Guoxin{discuss the result of model representation here.}

Our INF-3DP pipeline demonstrates strong flexibility in handling diverse model representations as inputs, including scanned point clouds (e.g., Statue-Goddess, Fig.~\ref{fig_statue}), B-rep parameterized surfaces (Satellite Bracket and TO-Connector, sampled along $u, v$ directions), and mesh vertices (graphics models). By leveraging SDF as a unified volumetric representation and applying a robust neural field learning process, our method effectively captures both complex topological and fine geometric features.
%- which significantly advance the accuracy in waypoint generation. This highlights the generalizability of our pipeline across a wide range of input modalities and model complexities.

\subsubsection{Waypoint accuracy and efficiency} With the integration of INF representation and the projection-based optimization for waypoint discretization presented in Sec.~\ref{sec_toolpath}, our approach achieves superior waypoint accuracy while maintaining excellent computational scalability. For example, the average point-to-surface error for the TO-Connector (input as parameterized surface, Fig.~\ref{teaser} is $0.019~mm$, which is less than $0.5\%$ compared with the waypoint spacing of $\Delta l = 2.5~mm$.
For comparison, waypoints generated using a dense mesh for TO-Connector (over $200k$ tetrahedra) with the mesh-based $S^{3}$-slicer~\cite{zhang2022s3} exhibit $12\times$ higher errors at $0.26~\text{mm}$ (see Fig.\ref{teaser}(g)). The computation time is also over 60 times slower than our method (826.4s \textit{v.s.} 12.9s) for models with a height of approximately $30\text{cm}$ (see Sec.~\ref{subsec:statistics} for details).
If the input of the model is represented by mesh and SDF training is based on vertices (e.g., the Light Bulb model shown in Fig.~\ref{fig:bubble}), we still achieve the average waypoint precision of $0.076~mm$, which still provides 63.46$\%$ improvement compare with fine-mesh-based result (average error is $0.208~mm$). %\Guoxin{data TBA.}

\subsubsection{Optimization of guidance field across computational domain} The INF representation enables the integration of fabrication-aware objectives for toolpath generation in both surface and interior domains. Through a two-stage optimization process for the guidance field, we achieve an vector field $\mathbf{v}_\text{T}$ on the model surface where singularities consistently converge to regions with minimal impact on surface quality (e.g., at the tips of branches in the Statue-Goddess model, as shown in Fig.~\ref{fig_statue}(c)). Optimizing the guidance field ensures excellent surface finish and support-free printability, greatly reducing the need for support structures in planar-based 3DP. Unlike previous methods~\cite{zhang2022s3, liu2024neural}, which optimize only in the 3-manifold domain without explicit singularity control on the surface, our approach delivers superior printing results (see Fig.~\ref{fig:scan}).
The comparisons of the support-free angle before and after optimization for the Fertility, Statue-Goddess, and Satellite Bracket models are shown in Fig.\ref{fig_collAblation}(b), Fig.\ref{fig_statue}(e), and Fig.~\ref{fig_collision}(g), respectively. The smoothness of the field also ensures that extrusion remains within practical limits for the extruder (see Fig.~\ref{fig_statue}(f)), for the printing setup with $D = 2.5~mm$ nozzle, the extrusion ratios are well controlled within $ [ 0.25D ,0.65D ] $. 

\begin{table*}[t]
\centering
\caption{Breakdown of computational times (in seconds) for each module in the INF-3DP pipeline.}
\label{tab:time}
\vspace{-5px}
\small % or \footnotesize or \scriptsize
\begin{tabular}{c c c || c || c c c || c c || c c c }
\hline
 & & Bond. Box & SDF & \multicolumn{3}{c||}{Guidance Fields Training} &  \multicolumn{2}{c||}{Waypoint Generation} & \multicolumn{3}{c}{Collision-Free Motion Planning}  \\
 \cline{5-7} \cline{8-9} \cline{10-12}
Model & Fig. & Size ($mm^3$) & Time & Surface$^{}$ & Interior & Infill & Waypoint $\#$ & Time & Sequence  & Motion & Total 
\\
\hline \hline 
TO-Connector      &   ~\ref{teaser}   &   $268 \times 235 \times 280$   &       404.88&     35.11&   6.03&    7.16&  2.46 M& 12.90&    17.72&  56.15 &   73.87\\
Bunny          &   ~\ref{fig_objective}   &   $200 \times 155 \times 194$   &   229.41&     21.23&   4.96&    5.06&  0.15 M&  2.09&    9.01 &  19.23 &  28.24 \\
Fertility         &   ~\ref{fig_pipeline}  &  $320 \times 125 \times 239$    &  483.20&     37.41&   6.62&    7.36&  1.25 M & 4.12&    16.81&   46.92&  63.73  \\
Satellite Bracket &    ~\ref{fig_collision}  &  $456 \times 131 \times 540$  &  687.17&     48.41&   7.17&    8.48&  1.34 M & 4.89&    42.58&  210.06&   252.64$^\dagger$  \\
Statue-Goddess    &   ~\ref{fig_statue}   &    $254 \times 153 \times 520$  &   590.45&     41.85&   5.97&    7.92&  0.21 M&  2.12&    23.06&   29.63&     52.69 \\
Light Bulb       &    ~\ref{fig:bubble}   &   $370 \times 370 \times 600$   &       731.52&     51.24&   8.21&    9.22&  1.74 M&  5.43&    32.47&   147.70&   180.17$^\dagger$    \\
\hline
\end{tabular}
\begin{flushleft}\footnotesize
$^\dagger$~Time includes iterative optimization, with partition refinement performed three times for Satellite Bracket and once for Light Bulb, respectively.
\end{flushleft}
\end{table*}

\begin{figure*}[t]
    \centering
    \includegraphics[width=1.0\linewidth]{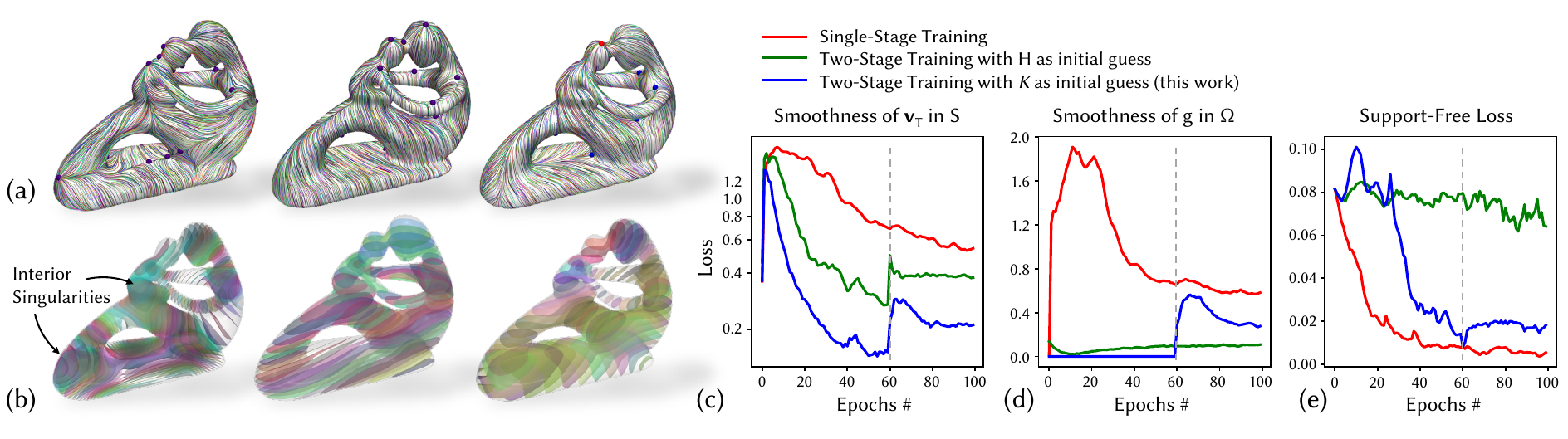}
    \caption{Training process and ablation study for the guidance field on the Fertility model. (a) Visualization of the vector field smoothness $\mathbf{v}_{\text{T}}$ on the two-manifold surface  $\mathcal{S}$; (b) iso-surfaces of $g$ on the volume three-manifold $\Omega$. From left to right: single-stage training of guidance INF, two-stage training with a height field as the initial guess, and warm-up training using a curvature field. (c) Learning curves for different loss terms: $\mathcal{L}_{\text{smooth}}^{\mathcal{S}}$, $\mathcal{L}_{\text{smooth}}^{\Omega}$, and $\mathcal{L}_{\text{SF}}^{\mathcal{S}}$.  Note that for two-stage training cases, the first 60 iterations employ $\mathcal{L}_{\text{guide}}^{\mathcal{S}}$ (Eq.~\ref{eq:guideSloss}), with subsequent iterations using $\mathcal{L}_{\text{guide}}^{\Omega}$ (Eq.~\ref{eq:guidePhiloss}) for each stages, respectively.}
    \label{fig_guideAblation}
\end{figure*}

As a general framework for both shell and solid structures, our neural field-based approach enables flexible and efficient generation of diverse infill patterns with optimized, singularity-free interior guidance fields. As demonstrated in Fig.~\ref{fig_infill}, we can easily realize a variety of infill structures — including cross, diamond, and hexagonal patterns — by training the infill fields $\psi$ effectively with different reference angles. For partial printing, we further demonstrate variable-density infills (see results in Fig.~\ref{teaser}(e) and Fig.~\ref{fig_infill}(b) for the TO-Connector and Bunny models, respectively). The variable-density cross infill adapts to the local density field derived from the SDF, which is not feasible with existing mesh-based solutions. 

\vspace{5px}
\subsubsection{Result of collision-free motion planning}~\label{subsec:motionPlanResult}
Our framework unifies the optimization of both the printing sequence and motion fields. By leveraging time-varying SDF evaluation and differentiable collision handling (as presented in Sec.~\ref{subsec:Time-SDF}), we achieve collision-free motion across all tested models. For the TO-Connector, Bunny, Fertility, and Statue-Goddess models, applying Reeb-graph-based partitioning ensures as-continuous-as-possible and collision-free printing motions (see Fig.~\ref{teaser}(e), Fig.~\ref{fig_objective}(c), Fig.~\ref{fig_pipeline}(e), Fig.~\ref{fig_statue}(d)). Notably, for the Statue-Goddess model, the nozzle performs bottom-to-top printing along the branches (illustrated in Fig.~\ref{fig_statue}(g)), highlighting the strong capability of our motion planning approach and the potential of multi-axis 3DP to achieve support-free fabrication.

For models with complex topology, such as the Light Bulb (Fig.~\ref{fig:bubble}) and Satellite Bracket (Fig.~\ref{fig_collision}), our iterative optimization strategy presented in Sec.~\ref{fig_collision} successfully achieves collision-free motion planning while maintain the motion continuity. We also show that the model size relative to the printing setup significantly affects the effectiveness of partition refinement.
For example, the Light Bulb model at heights of $200~\text{mm}$ and $600~\text{mm}$ (see Fig.\ref{fig:bubble}(d)) yields different printing sequences, as some branches cannot find feasible solutions with the initial partition (Fig.\ref{fig:bubble}(c)). Through iteration, our method consistently finds support-free and collision-free paths regardless of model complexity or waypoint count, with rapid convergence. With each iteration, fewer collision points need to be addressed, and reusing the motion INF leads to efficient convergence—requiring fewer than 10 epochs for the final two iterations of the 200~mm Light Bulb model (see Fig.~\ref{fig:bubble}(g)). Compared to existing point-based motion planning methods~\cite{dai2020planning, zhang2021singularity, chen2025co}, which require computationally intensive point-based global collision checking, our approach enables global collision detection via time-varying SDF (presented in Sec.~\ref{subsec:Time-SDF}) and enables continuous optimization on the motion INF; this further ensures the computational efficiency and scalability of our pipeline, as discussed below.

\subsection{Statistics and Scalability for Computation} 
\label{subsec:statistics}
Detailed computational statistics for each stage are provided in Table~\ref{tab:time}. Both guidance field optimization and motion planning are performed efficiently, with computation time largely influenced by model size. This is due to the physical-dimension-based sampling strategy discussed in~\ref{subsec:learningDetail}. The most time-consuming process is SDF training, which requires many iterations to minimize Laplacian loss and eliminate interior artifacts, as discussed in Sec.~\ref{subsec:SDF}. However, this step is essential for reliable infill generation and subsequent motion planning. 

We highlight the computational speed and scalability of our waypoint generation. Waypoint computation demonstrates strong linear scalability: generating 200k waypoints for the Bunny and Statue-Goddess models takes only a few seconds, and even the largest case (TO-Connector, 2.46M waypoints) requires less than 20 seconds. In contrast, mesh-based methods—relying on dense meshes, layer decomposition, and discrete waypoint tracing—are significantly less scalable (see Fig.~\ref{teaser}(h)). %For example, generating a similar number of waypoints with a nozzle diameter of 1.2~mm and waypoint spacing of 1.0~mm takes far longer time of for TO-Connector model (see Fig.\ref{teaser}). 
For the Light Bulb model, which features high-curvature regions and fine details, computation cost with mesh-based approaches will increase dramatically with higher mesh density, while our method efficiently generates 0.12 million waypoints for a 200~mm model (Fig.~\ref{fig:bubble}(e)) using 1.92s, and 5.43s for 1.74 million waypoints at 600~mm scale (Fig.~\ref{fig:bubble}(f)). %\Guoxin{Data TBA.}

%However, if a fine mesh is used for the computation, the toolpath generation time will easily exceed 300 seconds, while our method is 71 times faster than mesh-based solutions, requiring only 4.89 seconds to generate more than one million waypoints with GPU acceleration on Satellite Bracket model (detailed computation time discussed below).

%(for example, the precision enhancement is $0.56~mm$,  and $0.90~mm$ for TO-Connector and Light Bulb models, respectively (see comparison in Fig.\ref{teaser}(g) and Fig.~\ref{fig:bubble}(d)). 
%Note that if a fine mesh is used for the computation, the toolpath generation time will easily exceed 348.56 s, while our method is 71 times faster than mesh-based solutions, requiring only 4.89 seconds to generate more than one million waypoints with GPU acceleration on Satellite Bracket model (detailed computation time discussed below). %\Guoxin{To be finalized.}

%\Guoxin{discuss the result of guidance field training here.}

\subsection{Ablation Studies on Guidance Field Training}
\label{subsec:guideAblation}

When training the guidance field $g(x)$, it is necessary to jointly optimize field smoothness and compatibility across both surface and interior domains, along with support-free constraints. As discussed in Sec.~\ref{subsec:shortDiscuss}, directly putting all loss together will lead to poor convergence balance during training as these objectives can conflict with each other. To address this, we propose a two-stage training process to subsequently consider the loss defined in 2-manifold and 3-manifold domain (detail previously presented in Sec.~\ref{subsec:guidance}). 

The training process and ablation study of different strategies are demonstrated in Fig.~\ref{fig_guideAblation} with the Fertility model. With single-stage training, where all losses are optimized simultaneously, smoothness across the two domains is not well optimized, resulting in suboptimal field quality. While the support-free loss converges well due to the usage of compatibility loss $\mathcal{L}_{\text{align}}$ (Eq.~\ref{eq:align}) to enforces field merging at the boundary, it introduces significant variation in $\|\nabla g\|$ and singularities within the interior region that is not able to print (as seen in the left of Fig.~\ref{fig_guideAblation}(b)). In contrast, the two-stage training approach balances surface and interior field smoothness well. Our study also shows that the initial guess selection is crucial: if a height field is used (i.e., set $g(\textbf{p}) = z_{\textbf{p}}$ for warm-up training), the optimization tends to converge to a local minimum for $\mathcal{L}_{\text{smooth}}^{\Omega}$ where the internal iso-surfaces remain nearly parallel and unchanged (middle of Fig.~\ref{fig_guideAblation}(b)). Using the principal curvature field as the initial guess (Eq.~\ref{eq:warmup-guide}) secures good convergences for all objectives: surface singularities are minimized and consolidated in optimized places to ensure good surface finishing (shown in Fig.~\ref{fig_guideAblation}(a)), and the interior scalar field is smooth to ensure printable infill structure.

It is worth mentioning again that $\nabla g$ is not used as the final printing direction for each waypoint $\mathbf{d}_{\text{LPD}}$; therefore, the support-free loss does not necessarily have to converge to zero during this stage of training (right of Fig.~\ref{fig_guideAblation}(e)). Instead, we prioritize smoothness loss when training the guidance field $g(\mathbf{x})$ to enable the generation of high-quality toolpaths and infill patterns. The support-free constraints $\mathcal{C}_{\text{support}}$ (Eq.~\ref{eq_sf}) are further enforced during the subsequent training of the motion field (determined by the quaternion field $\mathbf{q}(\mathbf{x}))$, as discussed below.

\begin{figure}[t]
    \centering
    \includegraphics[width=1.0\linewidth]{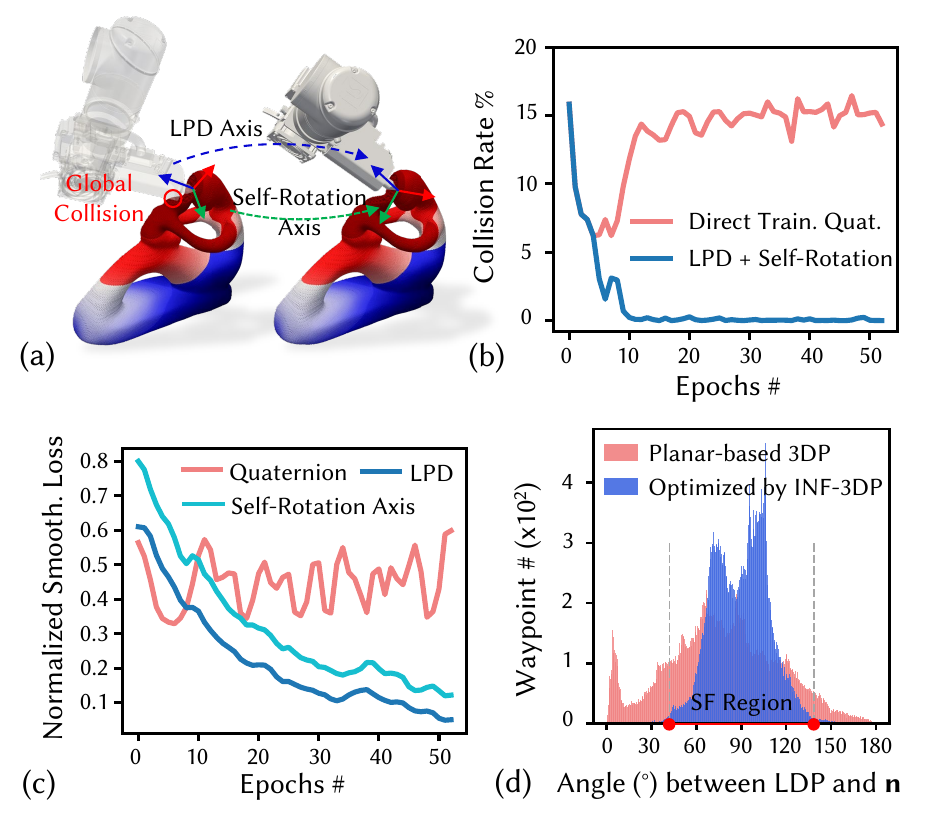}
    \caption{Ablation study of motion planning on the Fertility model. (a) Definition of LPD and self-rotation axis, optimized jointly with the quaternion INF. (b, c) Applying smoothness regularization directly to quaternions makes optimization hard to converge; this is resolved by defining smoothness loss separately for the LPD and self-rotation angle, as in Eq.~\ref{eq:smoothMotionLoss}. (d) The support-free constraint is satisfied with the optimized LPD.}
    \label{fig_collAblation}
\end{figure}

\subsection{Ablation Studies on Motion Field Training}

%\Guoxin{to discuss which model to demonstrate the frame field learning for the Quotation training.}

%\Js{We use frame fields with two cross axes, because the original guidance field $\mathcal{N}$ can offer the initial printing direction $\nabla \mathcal{N}$ compatible to the NN training. If use a vector field, such as Euler angles or quaternion, you have to warm up training then optimization of collision-free, which is non-efficient. Ablation studies are appended.}

To evaluate the effectiveness of our motion field training strategy, we conducted ablation studies on the Fertility model. As seen in Fig.~\ref{fig_collAblation}(a), the LPD is defined as the primary direction along which material is deposited, while the self-rotation axis specifies the local tool orientation required for continuous, collision-free fabrication. When optimizing for motion INF, directly including the quaternion field smoothness leads to a persistently high collision rate during motion planning (see Fig.~\ref{fig_collAblation}(b)). This failure is primarily due to the lack of explicit geometric constraints on the print direction, resulting in toolpaths that do not respect the physical requirements for support-free and collision-free printing, despite their good smoothness.  In contrast, we decouple the smoothness loss defined by the LPD and the self-rotation axis (Eq.~\ref{eq:smoothMotionLoss}), enabling robust convergence to collision-free motion (see blue curve in Fig.~\ref{fig_collAblation}(b)).

Fig.~\ref{fig_collAblation}(c) further highlights that optimizing the LPD and self-rotation axis yields lower smoothness loss compared to direct quaternion optimization, demonstrating better motion field regularity and printability. The support-free angle distribution presented in Fig.~\ref{fig_collAblation}(d) confirms that LPDs after training also satisfy the support-free constraint, ensuring successful printing of the Fertility model and demonstrating the practical benefits of our approach for complex geometries.

\subsection{Physical Experiments}

\begin{figure}[t]
    \centering
    \includegraphics[width=\linewidth]{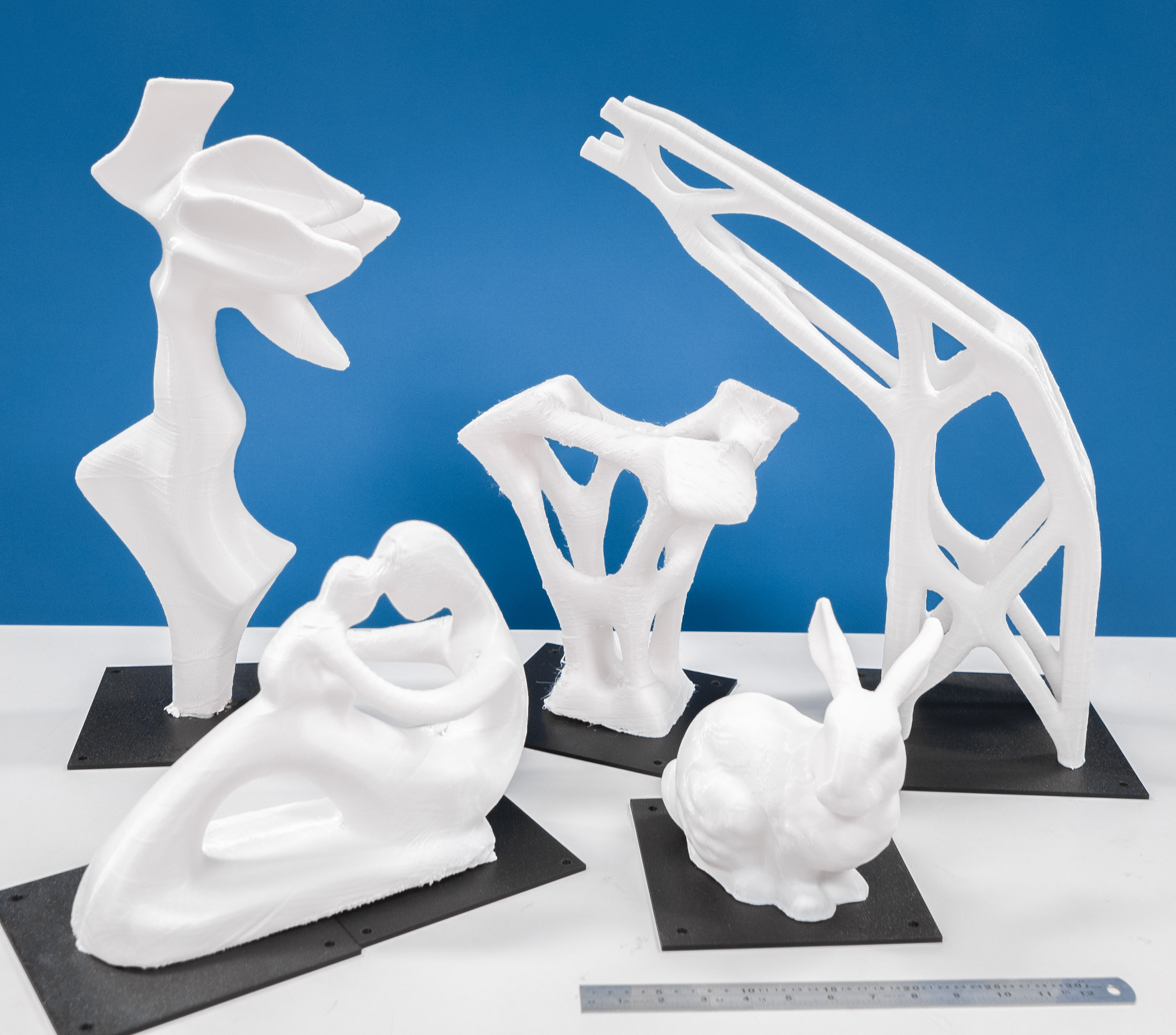}
    \caption{Result of physical fabrication. From left to right: Statue-Goddess, Fertility, TO-Connector, Bunny, and Satellite Bracket. All models are being printed by a robot-assisted multi-axis 3DP setup in a support-free manner with high surface finishing. The length of the ruler shown for reference is 300 mm.}
    \label{fig:printing}
\end{figure}

\begin{figure*}[t]
    \centering
    \includegraphics[width=\linewidth]{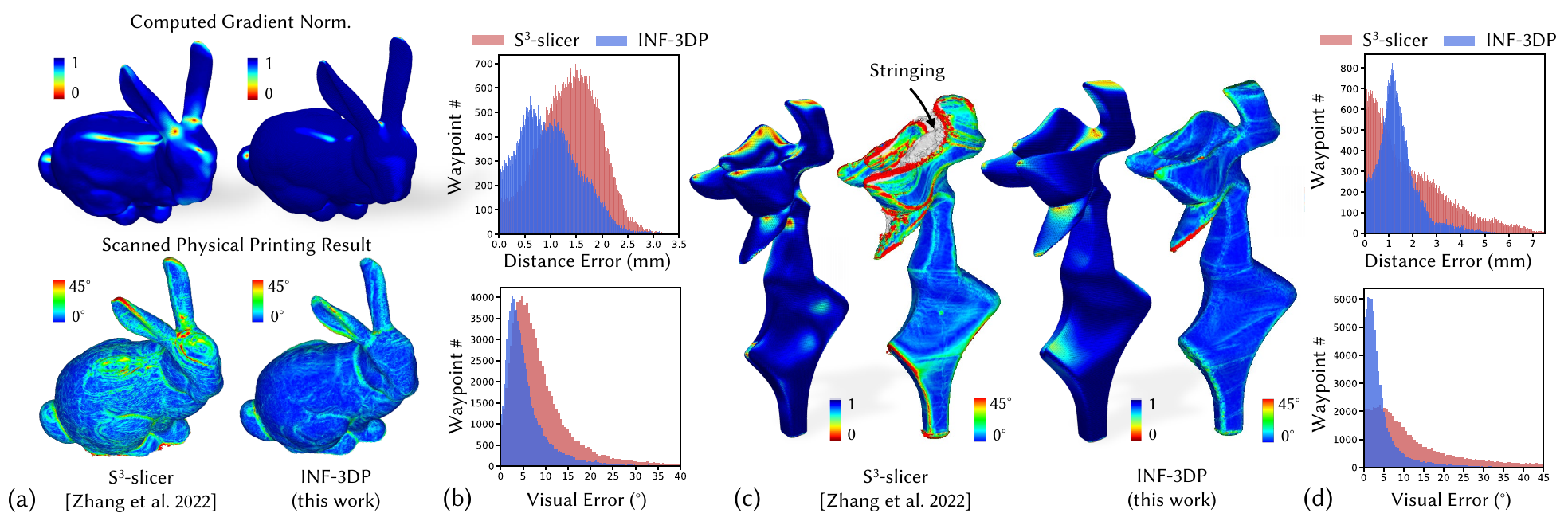}
    \caption{Comparison of fabrication errors with toolpaths generated by the baseline method~\cite{zhang2022s3}. (a) Existing tetrahedral mesh-based methods do not explicitly control singularity placement during field computation, resulting in significant printing artifacts. These artifacts, visualized by the norm of the projected guidance field gradient, highlight singularity convergence. (b) Our method achieves an error reduction of 28.47$\%$ in distance error (avg. error $0.98~mm$ \textit{v.s.} $1.37~mm$) and 39.10$\%$ in visual error (avg. error 5.98 $^\circ$ \textit{v.s.} 9.82 $^\circ$, computed as the difference in normal direction with the digital model). (d) Improvements are also observed for the Statue-Goddess model, with distance error reduced by 28.80$\%$ (avg. error $1.36~mm$ \textit{v.s.} $1.91~mm$) and visual normal error by 78.20$\%$ (avg. error $5.17^\circ$ \textit{v.s.} $23.88^\circ$), respectively. Note that severe stringing occurs at the top of the model with the baseline method due to its lack of continuity optimization, resulting in frequent large jumps in the toolpath.
 }
    \label{fig:scan}
\end{figure*}

We validated the computational results of INF-3DP through a series of physical fabrication experiments using a UR5e robotic arm equipped with a filament-based extruder (Dyze Design Typhoon™ High Flow Extruder) and PLA material (eSUN PLA+ 2.85~mm). Two nozzle sizes were used: 1.2~mm for models with intricate and delicate local features, minimizing the effect of heat accumulation (TO-Connector, Fertility, Satellite Bracket), and 2.5~mm for faster printing (Bunny, Statue-Goddess). As shown in Fig.~\ref{fig:printing} and Fig.~12, all test models were successfully fabricated, demonstrating advancements in optimal toolpath generation and collision-free motion planning. Further fabrication details are provided in the supplementary video.

To further demonstrate the superiority of our approach over existing solutions, we 3D scanned the fabricated Bunny and Statue-Goddess models to quantitatively evaluate fabrication errors. As shown in Fig.\ref{fig:scan}, our pipeline achieves a significant reduction in both distance and visual errors (evaluated by normal difference) compared to the mesh-based S$^3$-slicer~\cite{zhang2022s3}. For the Bunny model, our method reduces the average distance error by 28.47\% and the average visual error by 39.10\%. For the Statue-Goddess model, we observe a 28.80\% reduction in average distance error and a 78.20\% reduction in average visual error. The scanned results also show that the mesh-based method, which does not consider motion continuity during optimization, produces significant stringing artifacts. This can be seen at the top of the Statue-Goddess model in Fig.~\ref{fig:scan}(c). In contrast, our approach achieves high-quality surface finishing even for this challenging model that contains bottom-to-top printing direction (see Fig.~\ref{fig_statue}(g) and supplemental video for details).

% \Js{results to do list:
% \begin{enumerate}
%     \item Time statistics of each stage;
%     \item Spiral fish model computation
%     \item Statue model data. 1. scan image, SDF, streamlines, printing with robotarm down. 2. SF figure. 3. update fitting figure, with lower opacity, the sf figure floating on the figure.
%     \item Collision data with thick outer rectangle and larger ticks number.
%     \item Time-varying SDF and illustration. select a position making the Global SDF acceptable. Cut upper region of image.
%     \item Lattice infill with different patterns and density.
%     \item Scan of bunny. 
%     \item Pipeline image update, pcd, time-varying SDF, better nozzle pose.
%     \item update teaser line image color, ours is blue.
%     \item Animation of fields optimization, Time varying SDF.
% \end{enumerate}
% }

\subsection{Discussion and Limitation}
\label{subsec:limit}

We now discuss the advantages of the proposed INF-3DP over existing solutions as well as the limitations of this work.

\subsubsection{Advantages over Existing Solutions}

\begin{figure}[t]
    \centering
    \includegraphics[width=\linewidth]{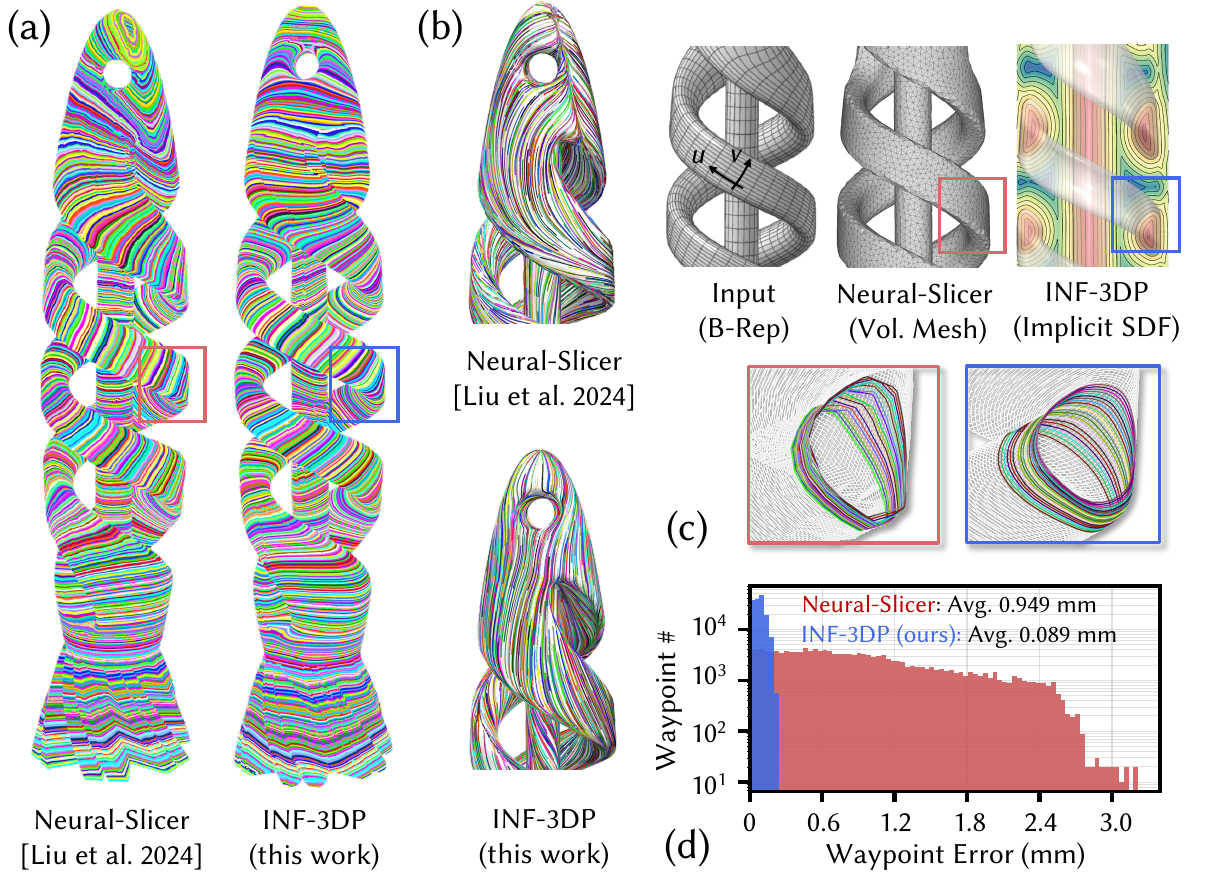}
    \caption{\rev{}{Comparison of performance with Neural-Slicer~\cite{liu2024neural} on the spiral fish model. (a) Both methods generate support-free toolpaths, but our method (b) achieves superior surface finishing by optimizing singularity placements. (c) With an input model represented by a parameterized surface, Neural-Slicer relies on a resolution-dependent mesh for computing, resulting in significantly higher waypoint errors, especially for highly curved regions. (d) Our method, by implicit field interpolation with INFs, reduces average waypoint errors to $9.3\%$ of Neural-Slicer’s. Additionally, our method presents a 20x faster computing speed: generating 140k waypoints in $1.46~s$ compared to Neural-Slicer's $29.4~s$ (data reproduced from~\cite{liu2024neural}).}}
    \label{fig:sprialFish}
%\Aoran{Is this an unreferenced figure? Fig. 13(c) looks outstanding. Being able to calculate Waypoint error implies we have a ground truth mesh, and we use a lower-resolution mesh for Nerual-Slicer, and we use the full-resolution mesh to train our nn SDF. Do we need to justify the resolution we used for Neural-Slicer? Such as this low-resolution mesh already takes longer time for Neural Slicer to converge than our method? }}
\end{figure}

The unified INF representation used in our framework offers several key advantages in computation. \rev{}{As demonstrated in Fig.~\ref{fig:sprialFish}, compared with most existing work of Neural-Slicer~\cite{liu2024neural},} the proposed method provides superior scalability and precision in toolpath computation, preserving fine geometric details in the generated waypoints that are often lost in mesh discretization. The implicit field-based approach also enables joint smoothness optimization across both surface and interior domains, ensuring more consistent and reliable toolpaths for both shell and solid printing. 

\rev{Our pipeline provides the solution to support differentiable motion planning, which allows for effective global collision checking through time-varying SDF. This is a critical factor for multi-axis 3DP and a common challenge for explicit waypoint-based methods. }{On the other hand, the unified INF representation used in our pipeline enables differentiable motion planning with efficient global collision checking via time-varying SDF, addressing a key challenge in multi-axis 3DP with explicit representation and waypoint-based motion planning methods. The TV-SDF reconstruction using INF interpolation ensures precise collision checking and extends to applications like accessibility analysis in subtractive manufacturing~\cite{zhong2025deepmill} and other 3DP processes~\cite{ariza2022adaptive}.}
%This is difficult to achieve with explicit representations that rely on point-wise collision checking.
%Overall, the INF-based framework achieves superior resolution and significantly higher computational efficiency, providing a general framework for multi-axis 3DP. 
%\Guoxin{discuss the advantage of time-varying SDF here - although the approximation error in collision distance, but the sign is guaranteed to be correct, support for effective optimization on the motion. - growing boundary }

\subsubsection{Configuration-Space Planning for Robot Arm}
One notable limitation of our current approach is its reliance on task-space planning rather than full configuration space planning~\cite{chen2025co}. In this work, the multi-axis printing setup is represented using only the final two joints of the robot arm and the extruder, which move together as a unit. This simplification reduces the complexity of collision checking involving other joints (e.g., the robot base) and lowers computational demands by omitting the influence of multi-solution inverse kinematics (IK) in the motion smoothness optimization~\cite{dai2020planning}. Although collisions with the robot base did not occur in any of our tested cases, since the base was positioned far from the model, integrating full inverse kinematics into the motion planning framework remains an important direction for future work.

% \subsubsection{Partition field training and iterative-based optimization}
% \Guoxin{consider remove this paragraph.} Another notable limitation arises in sequence field optimization, especially in regions where structural branches merge. In these areas, the confidence in sequence field assignments can be affected by training convergence, leading to ambiguity and potential suboptimal ordering of waypoints. While our method generally produces reliable results, complex branch merge scenarios may still introduce non-negligible risks of suboptimal sequencing. However, the overall impact on print quality and collision avoidance remains minor in our experiments. 

\subsubsection{Discussion on iteration-based partition and motion planning}
In our pipeline, the partitioning of the model to ensure continuous printing is optimized together with motion, which is iteratively refining the critical regions where collisions still being detected. Although this method works well in all tested cases and our top priority is to guarantee collision-free printing, it can lead to sub-optimal partitioning solutions that sacrifice continuity between regions. However, finding a global optimal solution by graph-based partition remain an open challenge due to the combinatorial complexity of the problem~\cite{huang2024learning,wang2025learning}. Developing more robust sequence field optimization strategies—potentially by incorporating global context or learning-based methods—represents an important direction for future work, with the potential to further improve the efficiency and reliability of automated toolpath planning for models with complex topologies.

%% file: tex/conclusion.tex
\section{Conclusion}

In this work, we present a novel computational pipeline for multi-axis 3D printing using INF representations to enable collision-free, high surface finishing, and collision-free fabrication. By unifying geometric, guidance, and motion fields within a differentiable neural framework, INF-3DP supports seamless optimization of fabrication objectives and global collision avoidance. Unlike mesh-based methods, our field-based approach overcomes resolution and scalability limits, leverages continuous toolpath and motion planning by INF optimization. Our iterative optimization of sequence and motion fields ensures globally collision-free and continuous printing. Extensive experiments demonstrate the generalizability, robustness, and superior performance of INF-3DP over traditional mesh-based solutions in terms of toolpath accuracy, global collision avoidance, and computational efficiency. INF-3DP provides a strong foundation for the next generation of intelligent, collision-free multi-axis 3D printing systems.